\documentclass[11pt]{article}

\usepackage[final]{acl}

\usepackage{times}
\usepackage{latexsym}

\usepackage[T1]{fontenc}

\usepackage[utf8]{inputenc}

\usepackage{microtype}

\usepackage{inconsolata}

\usepackage{graphicx}

\usepackage{amsmath,amsfonts,bm}

\def\eqref#1{equation~\ref{#1}}

\def\1{\bm{1}}

\def\vb{{\bm{b}}}

\def\vw{{\bm{w}}}
\def\vx{{\bm{x}}}

\def\mW{{\bm{W}}}

\DeclareMathAlphabet{\mathsfit}{\encodingdefault}{\sfdefault}{m}{sl}
\SetMathAlphabet{\mathsfit}{bold}{\encodingdefault}{\sfdefault}{bx}{n}
\newcommand{\tens}[1]{\bm{\mathsfit{#1}}}

\def\tC{{\tens{C}}}

\newcommand{\softmax}{\mathrm{softmax}}

\usepackage{xcolor}

\definecolor{aclblue}{rgb}{0, 0, 0.5}

\AtBeginDocument{
\hypersetup{
  colorlinks = true,
  linkcolor  = aclblue,
  citecolor  = aclblue,
  urlcolor   = aclblue
}}

\usepackage{wrapfig}
\usepackage{subcaption}
\usepackage{enumitem}
\setlist[enumerate]{leftmargin=2em}
\usepackage{booktabs}
\usepackage{multirow}
\usepackage{float}

\definecolor{cgpt}{HTML}{00E079}
\definecolor{cclaude}{HTML}{9B4400}
\definecolor{lightblue}{HTML}{00adef}
\definecolor{darkblue}{HTML}{0079b7}
\definecolor{cfp}{HTML}{BF1E2E}
\definecolor{cff}{HTML}{136783} 
\definecolor{upperboundc}{HTML}{BF1E2E} 
\definecolor{correct}{HTML}{FF7777} 
\definecolor{incorrect}{HTML}{69c39a} 
\definecolor{numbery}{HTML}{5BA8EA} 
\newcommand{\bmf}[1]{\bm{\mathsf{#1}}}
\newcommand{\darkblue}[1]{\textcolor{darkblue}{#1}}
\newcommand{\lightblue}[1]{\textcolor{lightblue}{#1}}
\newcommand{\black}[1]{\textcolor{black}{#1}}
\newcommand{\hl}{\darkblue{\bmf{h}_1}}
\newcommand{\ho}{\lightblue{\bmf{h}_0}}
\newcommand{\h}{\black{\bmf{h}}} 
\newcommand{\z}{\black{\bmf{z}}}

\newcommand{\correct}[1]{\textcolor{correct}{#1}}
\newcommand{\incorrect}[1]{\textcolor{incorrect}{#1}}
\newcommand{\numbery}[1]{\textcolor{numbery}{#1}}
\newcommand{\yc}{\correct{y_c}}
\newcommand{\yi}{\incorrect{y_i}}
\newcommand{\yn}{\numbery{y_n}}
\newcommand{\cn}{\char`\\n}
\def\vx{{\bm{x}}}
\def\vw{{\bm{w}}}
\def\mW{{\bm{W}}}
\def\vb{{\bm{b}}}

\def\tC{{\tens{C}}}
\def\deltaLogP{{\Delta \log \pi (y_t\mid\h)}}
\def\deltaH{{\Delta \h}}
\def\gradient{{\nabla_{\h} \log \pi (y_t|\h_0)}}
\def\upperBound{{\|\gradient\| \cdot \|\deltaH\|}}

\title{Understanding the Prompt Sensitivity}

\author{Yang Liu \quad  Chenhui Chu \\
        Kyoto University \\
        \texttt{yangliu@nlp.ist.i.kyoto-u.ac.jp}, \texttt{chu@i.kyoto-u.ac.jp}}

\begin{document}
\maketitle
\begin{abstract}
Prompt sensitivity, which refers to how strongly the output of a large language model (LLM) depends on the exact wording of its input prompt, raises concerns among users about the LLM's stability and reliability.
In this work, we consider LLMs as multivariate functions and perform a first-order Taylor expansion, thereby analyzing the relationship between meaning-preserving prompts, their gradients, and the log probabilities of the model's next token. 
We derive an upper bound on the difference between log probabilities using the Cauchy-Schwarz inequality.
We show that LLMs do not internally cluster similar inputs like smaller neural networks do, but instead disperse them.
This dispersing behavior leads to an excessively high upper bound on the difference of log probabilities between two meaning-preserving prompts, making it difficult to effectively reduce to 0. 
In our analysis, we also show which types of meaning-preserving prompt variants are more likely to introduce prompt sensitivity risks in LLMs.
In addition, we demonstrate that the upper bound is strongly correlated with an existing prompt sensitivity metric, \texttt{PromptSensiScore}.
Moreover, by analyzing the logit variance, we find that prompt templates typically exert a greater influence on logits than the questions themselves. 
Overall, our results provide a general interpretation for why current LLMs can be highly sensitive to prompts with the same meaning, offering crucial evidence for understanding the prompt sensitivity of LLMs.
Code for experiments is available at \url{https://github.com/ku-nlp/Understanding_the_Prompt_Sensitivity}.
\end{abstract}

\section{Introduction}
Large language models (LLMs) usually show sensitivity to even minor variations in prompts, such as wording, prompt template, or even minor spelling errors, although these variations do not change the meaning of the prompt~\citep{chatterjee-etal-2024-posix}.
This phenomenon can be described as LLMs' prompt sensitivity, which can amplify the output variance, making the model's output unreliable. 
To quantify this effect, researchers~\citep[]{zhuo-etal-2024-prosa,chatterjee-etal-2024-posix} have made considerable efforts to assess the sensitivity of LLMs to minor variations in prompts.
Also, \citet[]{sun2024evaluating} have attempted to improve the generalization ability of LLMs through reinforcement learning from human feedback~\citep[RLHF;][]{christiano2017deep} or instruction tuning~\citep{wei2021finetuned}.
However, even minor changes such as prompt formatting to the wording of the prompts still can lead to the prompt sensitivity of these models~\citep{sclar2024quantifying}.

Although prompt sensitivity in LLMs is frequently highlighted, its generation mechanism remains poorly understood.
For example, we still do not understand why a set of meaning-preserving prompts can yield completely different outputs by an LLM.
This open issue leads to a lack of credibility in previous benchmark-based prompt sensitivity evaluations~\citep{zhuo-etal-2024-prosa, chatterjee-etal-2024-posix} and the arbitrary practice of fine-tuning LLMs by increasing training samples~\citep{liu-etal-2025-take,dong-etal-2024-abilities}. Previous studies~\citep{zhuo-etal-2024-prosa, chatterjee-etal-2024-posix} calculate a metric to represent a model's sensitivity to wording changes in prompts based on its output.
However, they make only limited contributions to understanding the prompt sensitivity of LLMs and fail to guide fundamental breakthroughs.

Contrary to previous studies, we aim to understand the prompt sensitivity of LLMs using a mathematical analysis method: Taylor expansion~\citep{taylor1715methodus}. 
In this study, we focus on transformer-based LLMs.
Specifically, we formalize an LLM as a continuous multivariate function that outputs the log probability of the model's next token. 
The hidden states are responsible for converting the prompts in discrete space into the continuous representation space. 
Then, we use the first-order Taylor expansion of this function to connect the hidden states of the prompt with the output log probabilities. 
Furthermore, we monitor the changes in hidden states of two meaning-preserving prompts' across layers to explain the prompt sensitivity of LLMs.

Our analysis starts with an image classification task. We observe that ResNet~\citep{he2016deep} internally produces clustering behavior to achieve high classification accuracy.
Next, we build connections between two prompts using Taylor expansion, derive an upper bound for the log probability difference via the Cauchy-Schwarz~\citep{cauchy1821cours,schwarz1890ueber} inequality, and reveal why LLMs exhibit prompt sensitivity by observing their different behaviors compared to traditional neural networks (RQ1). 
We then investigate which types of prompt modifications are more likely to lead to prompt sensitivity (RQ2). 
We also find that the upper bound correlates strongly with an existing prompt sensitivity metric (RQ3). 
Furthermore, by analyzing the variance of LLMs' logits, we observe that in existing LLMs, prompt templates exert a greater influence on logits than the questions themselves~(RQ4).

\section{Neural Networks Are Functions}
\label{sec:cifar10_example}

A neural network is a mathematical relationship that maps inputs to outputs~\citep{lecun2015deep,nielsen2015neural,goodfellow2016deep}. 
If the input is a vector $\vx \in \mathbb{R}^d$ and the output is a scalar $y \in \mathbb{R}$, then a single-layer neural network can be represented as:
\begin{equation}
    y = \sigma(\vw^\top \vx + b)
\end{equation}
where $\vw$ is the weight vector, $b$ is the bias, and $\sigma(\cdot)$ is the activation function, such as sigmoid~\citep{rumelhart1986learning}, ReLU~\citep{nair2010rectified, glorot2011deep}, etc.
If we consider this neural network as a function $y = f(\vx)$. The one-time inference using this neural network can be interpreted as input vector $\vx$ to the function $f(\vx)$, outputting the scalar $y$.
In this section, we start by explaining why deep neural networks are composite functions. Then, we introduce intra-class mean distances, a simple representation of space distances. Finally, we interpret why deep neural networks can perform classification tasks from an interesting perspective: that a neural network is a function.

\paragraph{Deep neural networks are compositions of functions.} 


A deep neural network defines a function as a composition of simpler functions.
In particular, it is composed of layer-by-layer composites of affine transformations and activation functions~\citep{cybenko1989approximation,hornik1989multilayer,murphy2012machine}.
Formally, the affine transformation of the layer $l$ is:
\begin{equation}
    A_l(\vx) = \mW_l \vx + \vb_l
\end{equation}
where $\vx \in \mathbb{R}^{d_{l-1}}$ is the output vector of layer $l-1$, $\mW_l \in \mathbb{R}^{d_l \times d_{l-1}}$ is the weight of the layer $l$, and $\vb$ is the bias of layer $l$.
Then, the affine transformation $A_l(\vx)$ is composed using the activation function $\sigma_l $ as follows:
\begin{equation}
    g_l = \sigma_l \circ A_l
\end{equation}
The general mapping of the deep neural network of layer $L$ is as follows:
\begin{equation}
    F = g_L \circ g_{L-1} \circ \cdots \circ g_1
\end{equation}
where the composite function $F$ is a continuous mapping from the input space to the output space~\citep{goodfellow2016deep}.


\paragraph{Intra-class compactness.} Intra-class compactness~\citep{yanluo2020g} refers to how close or tightly clustered the samples or data points of the same class are in the feature space. Typically, an ideal classifier requires ensuring high intra-class compactness. 
To remove the influence of vector dimension on distance metrics, we first perform $L^2$ normalization on the feature vectors, then use the Euclidean distance between the normalized vectors as the metric:
\begin{equation}
\label{eq:5}
    d(\vx_i,\vx_j) = \|\vx_i-\vx_j\|
\end{equation}
As all vectors are normalized to the unit hypersphere, this distance reflects only directional differences and is equivalent to cosine similarity, e.g., $\|\vx_i-\vx_j\| = \sqrt{2-2\cos \theta_{ij}}$ (see Appendix~\ref{sec:proof}), where $\theta_{ij}$ is the angle between the two vectors. We denote the samples for class $c$ as $\mathcal{J}_c$. We use the intra-class mean distance as the metric. The distance of samples in class $c$ is defined as follows:
\begin{equation}
    D_{\mathrm{intra}}^{(c)} = \frac{1}{|\mathcal{J}_c| (|\mathcal{J}_c| - 1)} \sum_{\vx_i , \vx_j \in \mathcal{J}_c, i\ne j}^{} d(\vx_i,\vx_j)
\end{equation}


We use the average of the distances over all classes as the metric for intra-class compactness:
\begin{equation}
    D_{\mathrm{intra}} = \frac{1}{|\mathcal{C}|}\sum_{c \in \mathcal{C}} D_{\mathrm{intra}}^{(c)}
\end{equation}
where $\mathcal{C}$ is the class set and $|\mathcal{C}|$ denotes the total number of classes. Enhancing intra-class compactness, i.e., decreasing $D_{\mathrm{intra}}$, can improve the neural network's classification performance~\citep{10.5555/3045390.3045445,yanluo2020g}.

\paragraph{Intra-class mean distance of ResNet on CIFAR-10.}

\begin{figure}[t]
  \includegraphics[width=\columnwidth]{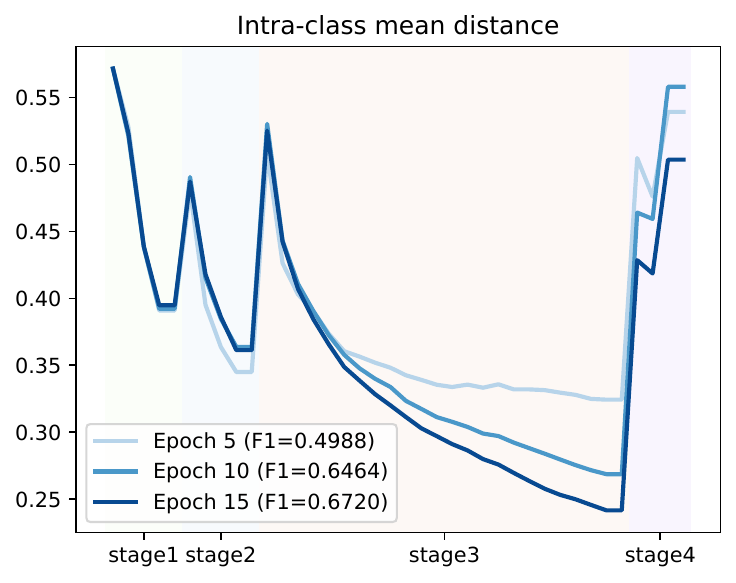}
  \caption{Intra-class mean distances for CIFAR-10 across different training epochs.}
  \label{fig:intra_class_mean_distance}
\end{figure}

To illustrate the internal behavior of neural networks while performing classification tasks, we investigate how the intra-class compactness of the feature maps changes across each stage of the neural network. 
We take the application of ResNet~\citep[we pick \texttt{ResNet-101;}][]{he2016deep} on the CIFAR-10 dataset~\citep{krizhevsky2009learning} as an example. ResNet is typically divided into four stages, each consisting of multiple residual blocks stacked together and outputting the feature maps of the input image. 
Stages 1, 2, 3, and 4 consist of 3, 4, 23, and 3 blocks, respectively.
Here, we focus on the block level of ResNet to analyze the features output by each block and the input features output by the stem module of ResNet.

We trained this neural network on the CIFAR-10 dataset for 20 epochs, achieving the highest F1 score of 0.7048 on the testing set at epoch 18.\footnote{For more hyperparameters, see Appendix~\ref{sec:the_hyperparameters_for_training_resnet}.}
As shown in Figure~\ref{fig:intra_class_mean_distance}, we compare the intra-class mean distance of features across three epochs: epoch 5, 10, and 15; their F1 scores gradually increased.
A low intra-class mean distance indicates high intra-class compactness. 
At the stage level, we observe that the intra-class mean distance gradually decreases from stage 1 to stage 3, then increases in stage 4.
This indicates that stages 1 to 3 are performing clustering, while stage 4 is classifying feature differences within classes.
The behavior of classifying feature differences within classes also occurs in stages 2 and 3.
In stage 3, which demonstrated the best clustering performance, epochs with lower intra-class mean distances yield higher F1 scores. 
This indicates that \textbf{the neural network achieves stronger classification performance by improving its clustering behavior}.
From the functional perspective, clustering brings samples of the same class closer together, while continuous functions produce similar outputs for similar inputs.
In this paper, based on the above analysis, we formulate LLMs as functions and investigate their prompt sensitivity using Taylor expansion.

\section{Interpretation of Prompt Sensitivity}
\label{sec:interpretation_of_prompt_sensitivity}
The prompt sensitivity of LLMs usually refers to minor variations in prompts causing LLMs to respond with different results~\citep{zhuo-etal-2024-prosa,chatterjee-etal-2024-posix}.
In this paper, we narrow it down to describe ``how prompt $p_0$ and its meaning similar prompt $p_1$ cause the LLMs to respond with different log probabilities of the model's next token $y_t$.'' 
The natural language prompts or their tokenized tokens reside in a discrete space, while their embeddings represented by the embedding layer or hidden states output by a specific transformer block can be regarded as variables in the continuous representation space.

\subsection{LLMs Are Multivariable Functions}
\label{sec:llm_function}

In \S~\ref{sec:cifar10_example}, we observe that ResNet exhibits clustering behavior to achieve significantly higher accuracy and F1 score.
In this section, we generalize this interpretation to LLMs. 
Unlike classification neural networks, which project the feature representations into a class space, LLMs project the feature representations into a vocabulary space to predict the next token~\cite{vaswani2017attention}.

In LLMs' inference stage, when an LLM predicts the next token, it first maps the input tokens into embeddings by the embedding layer and adds positional encodings to form a sequence representation. Then, the sequence representation passes through several transformer blocks sequentially. In the self-attention module of each transformer block, a causal mask is applied to block tokens to the right of the current position, ensuring that each current position only depends on the content to its left. In this way, each position ultimately obtains a hidden state vector that contains only the prefix information.
When the model processes the entire input sequence, it predicts the next token using the hidden state of the last position.
This hidden state is projected to the vocabulary space via the output layer (typically a linear layer and softmax).

Now, suppose we input a prompt containing $L$ tokens into an LLM. 
The model maps each token in this prompt to a $D$-dimensional embedding. 
Following the analysis in \S~\ref{sec:cifar10_example}, 
We consider an LLM as a multivariable function, where the output of the embedding layer serves as the function's input. The log probability of the next token is treated as the model's output value.
The difference between the log probabilities of two meaning-preserving inputs (prompts) can be interpreted as a measure of prompt sensitivity. 
A smaller difference indicates lower prompt sensitivity of the model.

\subsection{Taylor Expansion of LLMs}
\label{sec:taylor_expansion_of_llms}

We use the Taylor expansion to build connections between two meaning-preserving prompts.
For simplicity, we denote the log probability difference between two meaning-preserving prompts $\ho$ and $\hl$ as $\deltaLogP$.
Here,
\begin{equation}
    \deltaLogP = \log \pi (y_t\mid\hl ) - \log \pi (y_t\mid\ho),
\end{equation}
where $\pi = \softmax(\z)$ is the softmax of the logits $\z$ output by the LLM.
Formally, we can express the relationship between the hidden states $\ho$ and $\hl$ by Taylor expansion\footnote{Appendix~\ref{sec:taylor_expansion} provides the first-order Taylor expansion for both univariate and multivariate cases.} as follows:
\begin{equation}
\begin{aligned}
\underbrace{\deltaLogP}_{1\times 1}
&= \underbrace{\nabla_{\h}\log \pi (y_t\mid\ho)^\top}_{1\times D} \underbrace{(\deltaH)}_{D \times 1} \\
&+ \mathcal{O}(\|\deltaH\|^2),
\end{aligned}
\label{eq:tailor_llm}
\end{equation}
where $\mathcal{O}(\|\deltaH\|^2)$ is the remainder term of the Taylor expansion. $\nabla_{\h} \log \pi(y_t\mid\ho)$ indicates the gradient vector and $\deltaH = \hl -\ho$ indicates the difference between the feature representations of the two prompts.
It is calculated through element-wise subtraction, thus capturing not only semantic differences between the two prompts but also variations in their expressive styles. In this paper, unless otherwise specified, we set the correct answer of the question as $y_t$. Discussions regarding other tokens as $y_t$ are provided in Appendix~\ref{sec:tailor_llm_more_tokens}.

\subsection{Upper Bound}
From the properties of the Taylor expansion, we know that when the distance between $\ho$ and $\hl$ is sufficiently close, the remainder term $\mathcal{O}(\|\deltaH\|^2)$ will vanish faster than $\|\deltaH\|^2$ as $\hl \to \ho$.
Moreover, in this paper, $\h_0$ and $\h_1$ are two meaning-preserving prompt words that reside in a close semantic space.
Based on this condition, we rewrite Eq.~(\ref{eq:tailor_llm}) in the following form:
\begin{equation}
    \deltaLogP \approx \nabla_{\h} \log \pi(y_t\mid\h_0)^\top \deltaH.
    \label{eq:tailor_approximate}
\end{equation}
Then, we obtain the following inequality by calculating the L2 norm:
\begin{equation}
    |\deltaLogP|
\le \upperBound,
\label{eq:ineq}
\end{equation}
where $\|\cdot\|$ is the L2 norm.
This inequality tells us that $|\deltaLogP|$ has an upper bound $\upperBound$. 
If the upper bound is significantly low, $|\deltaLogP|$ can be approximated as 0, meaning the two meaning-preserving prompts receive equal log probabilities of the model's next token.

\paragraph{Calculate the gradient.}
We represent the gradient matrix as follows:
\begin{equation}
    \nabla_{\h} \log \pi(y_t\mid\h_0)^\top = G(\h_0),
\end{equation}
where $G(\h_0) \in \mathbb{R}^{D}$ represents the gradient vector of $\h_0$. 
The gradient for the $i$-th dimension is calculated as follows:
\begin{equation}
    g(\h[i]) = \nabla_{\h[i]}\log \pi(y_t\mid\h)
\end{equation}
The gradient $g(\h[i])$ is usually named the saliency score~\citep{simonyan2013deep, li2016visualizing}.
Unlike~\citet{yin-neubig-2022-interpreting}, who use the L1 norm to calculate the saliency score for each input token, we take the L2 norm of the gradient vector to obtain the saliency score of the input $\h$ as follows:
\begin{equation}
    S_{GN}(\h) = \|\nabla_{\h} \log \pi(y_t\mid\h)\|_2 = \sqrt{\sum_{i} \left| g(\h[i])\right|^2 } 
\end{equation}
$S_{GN}(\h)$ is the overall contribution of $\h$ to the log probability of the model's next token.

\section{Experimental Verifications}
\label{sec:experimental_verifications}

\begin{figure}[t]
    \centering
    \begin{subfigure}[b]{0.37\textwidth}
        \centering
        \includegraphics[width=\linewidth]{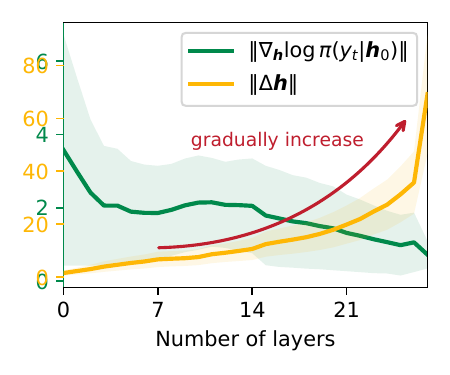}
        \caption{The trend of $\|\gradient\|$ and $\|\deltaH\|$.}
        \label{fig:result_of_rq1_1}
    \end{subfigure}
    \begin{subfigure}[b]{0.37\textwidth}
        \centering
        \includegraphics[width=\linewidth]{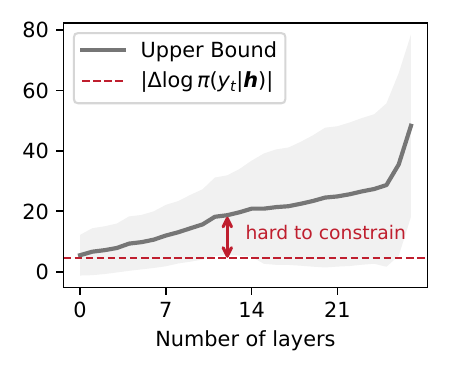}
        \caption{The trend of the upper bound.}
        \label{fig:result_of_rq1_2}
    \end{subfigure}
    
    \caption{
    Key results of RQ1: (a) indicates the trend of $\|\gradient\|$ and $\|\deltaH\|$ across layers. (b) indicates the trend of the upper bound across layers. The upper bound is calculated by multiplying $\|\gradient\|$ by $\|\deltaH\|$.
    }
    \label{fig:result_of_rq1}
\end{figure}

In this section, we verify our analytical results in practical settings.
We consider four multiple-choice question (MCQ) datasets commonly used to evaluate prompt sensitivity~\citep{zhuo-etal-2024-prosa,chatterjee-etal-2024-posix}, ARC Challenge~\citep{clark2018think}, CommonSenseQA~\citep{talmor-etal-2019-commonsenseqa}, MMLU~\citep{hendrycks2021measuring}, and OpenBookQA~\citep{OpenBookQA2018}. 
To further examine whether our analysis generalizes beyond MCQ tasks, we additionally include the open-ended generation dataset Alpaca~\cite{alpaca}.
Each MCQ sample is a multiple-choice question with a correct option as the target token, and each Alpaca sample consists of an instruction, where the last token of its reference response is taken as the target token. 
We randomly select 500 examples to create our test set from each dataset.
We consider 12 prompt templates\footnote{All templates are provided in Appendix~\ref{sec:meaning-preserving_prompts}.} proposed by~\citet{zhuo-etal-2024-prosa} as the meaning-preserving prompts for LLMs.
We combine the 12 prompt templates with 500 samples from each dataset.
We perform all our experiments on four model series: \texttt{Pythia-410M/1B/1.4B}~\citep{biderman2023pythia}, \texttt{GPT2-small/medium/large}~\citep{radford2019language}, \texttt{Qwen1.5-0.5B/1.8B/4B}~\citep{bai2023qwen}, and \texttt{Llama3.2-1B/3B}~\citep{touvron2023llama}. 

\subsection{Why Do LLMs Exhibit Prompt Sensitivity? (RQ1)}
In this section, we explain why LLMs exhibit prompt sensitivity.
Specifically, we observe the trend of the upper bound ($\upperBound$) in Eq.~(\ref{eq:ineq}) across LLM layers.
As shown in Figure~\ref{fig:result_of_rq1_1}, we illustrate the trends of $\|\gradient\|$ (the green line) and $\|\deltaH\|$ (the yellow line) across the layers of \texttt{Llama3.2-3B} on the ARC Challenge dataset.\footnote{Experimental results for other models are provided in Appendix~\ref{sec:more_results_of_rq1}.} We observe that the gradient values are higher in the earlier layers of the model and lower in the later layers.
However, unlike the clustering behavior observed in traditional classification tasks, $\|\deltaH\|$ \textbf{gradually increases} from close to 0 to approximately 70 across the model layers.
Because the upper bound is calculated by multiplying $\|\gradient\|$ by $\|\deltaH\|$. Additionally, the gradient is not less than 0 (Figure~\ref{fig:result_of_rq1_2}).
The increase of $\|\deltaH\|$ leads to an increasing trend of the upper bound across layers, making it impossible to converge to sufficiently low values and \textbf{hard to constrain} $|\deltaLogP|$ to 0 via the upper bound.
This means that the log probabilities of the next token for the two meaning-preserving prompts are not exactly equal.

In summary, we have the following interpretation for prompt sensitivity of LLMs. First, LLMs do not exhibit the clustering behavior that is found in traditional neural networks. 
This clustering behavior serves as a crucial role in allowing neural networks to accurately perform classification tasks.
Secondly, as LLMs tend to pull meaning-preserving prompts farther apart in the representation space, this leads to giving $|\deltaLogP|$ a large upper bound $\upperBound$. This makes it hard to constrain $|\deltaLogP|$ to 0.
In other words, because LLMs do not exhibit clustering behavior for meaning-preserving prompts, they can only learn each sample individually during training. This cannot guaranty that the model fits meaning-preserving prompts to the same degree, leading to different outputs.
To further verify the causal role of $\|\deltaH\|$, we conduct an activation steering experiment that directly forces $\|\deltaH^{(l)}\| = 0$ at a chosen layer; steering consistently reduces the observed prompt sensitivity, empirically confirming this causal role. Details and full results are provided in Appendix~\ref{sec:mitigating}.

\subsection{Which Types of Prompt Modifications Are More Likely to Lead to Higher Upper Bounds? (RQ2)}
\label{sec:real-world_dataset_validation}

\begin{figure*}[t]
    \centering
    \begin{subfigure}[b]{0.312\textwidth}
        \centering
        \includegraphics[width=\linewidth]{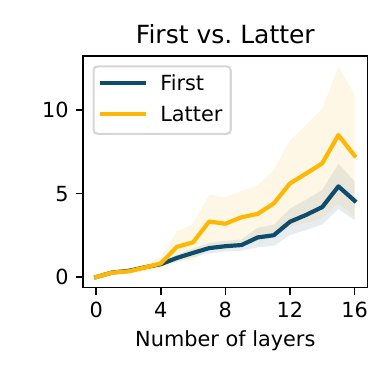}
        \caption{Pythia-1B.}
        \label{fig:result_of_rq2_1}
    \end{subfigure}
    \begin{subfigure}[b]{0.30\textwidth}
        \centering
        \includegraphics[width=\linewidth]{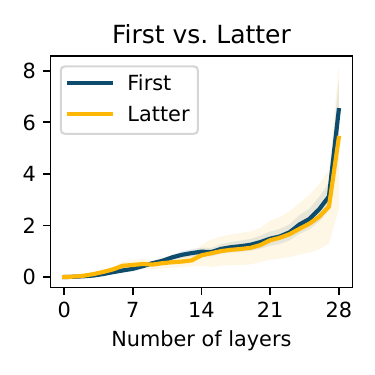}
        \caption{Llama3.2-3B.}
        \label{fig:result_of_rq2_2}
    \end{subfigure}
    \begin{subfigure}[b]{0.30\textwidth}
        \centering
        \includegraphics[width=\linewidth]{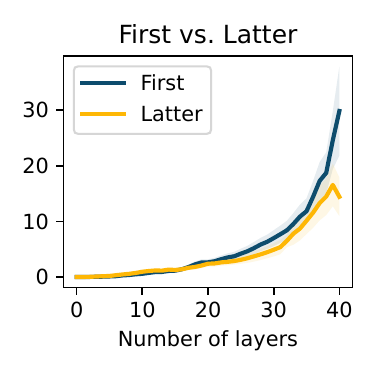}
        \caption{Qwen1.5-4B.}
        \label{fig:result_of_rq2_3}
    \end{subfigure}
    \hfill
    \begin{subfigure}[b]{0.30\textwidth}
        \centering
        \includegraphics[width=\linewidth]{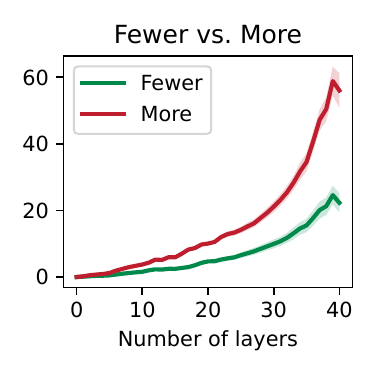}
        \caption{Misalignment.}
        \label{fig:result_of_rq2_4}
    \end{subfigure}
    \begin{subfigure}[b]{0.30\textwidth}
        \centering
        \includegraphics[width=\linewidth]{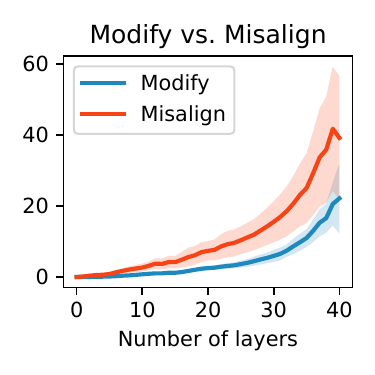}
        \caption{Modification vs. Misalignment.}
        \label{fig:result_of_rq2_5}
    \end{subfigure}
    \begin{subfigure}[b]{0.30\textwidth}
        \centering
        \includegraphics[width=\linewidth]{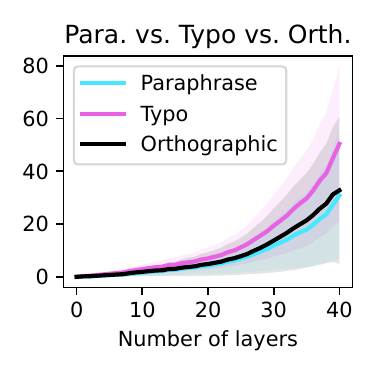}
        \caption{Para. vs. Typo vs. Orth.}
        \label{fig:result_of_rq2_6}
    \end{subfigure}

    \caption{Key results of RQ2: (a), (b), and (c) indicate the trend of $\|\deltaH\|$ when modifying the first and latter half of the prompt templates.
    (d), (e), and (f) are results on \texttt{Qwen1.5-4B}; (d) and (e) use the ARC Challenge dataset, (f) uses the Alpaca dataset.
    (d) indicates the trend of $\|\deltaH\|$ when the prompt templates have fewer and more tokens misaligned. 
    (e) indicates the trend of $\|\deltaH\|$ between modification and misalignment.
    (f) indicates the trend of $\|\deltaH\|$ among paraphrase, typo, and orthographic modifications.
    Full results are provided in Appendix~\ref{sec:more_results_of_rq2}}
    \label{fig:result_of_rq2}
\end{figure*}

\begin{figure*}[t]
\centering
  \includegraphics[width=\linewidth]{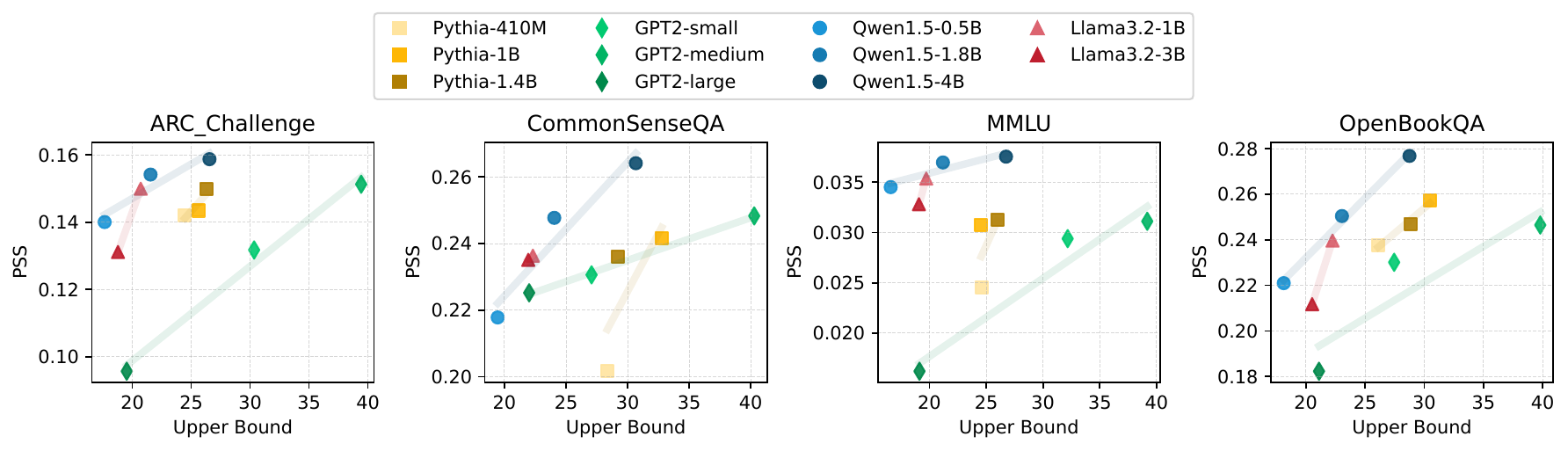}
  \caption{The relationship between the average upper bound across layers of LLMs and PSS. The lines represent linear fits of points within the same model series.}
  \label{fig:result_of_rq3}
\end{figure*}

Understanding which types of prompt modifications are more likely to lead to prompt sensitivity is important.
This can provide evidence for anticipating and preventing biases caused by prompt sensitivity.
To investigate which types of prompt modifications are more likely to lead to higher prompt sensitivity, we create seven modification types to modify the prompt template.
Our seven modification types are as follows:\footnote{For details of the prompt templates, please refer to Appendix~\ref{sec:modification_and_misalignment_prompt_templates}.}

\begin{enumerate}
    \item Modification first ($\h_{first}$): replace one token in the first half of the seed prompt template with a synonymous token.
    \item Modification latter ($\h_{latter}$): replace one token in the latter half of the seed prompt template with a synonymous token.
    \item Misalignment fewer ($\h_{fewer}$): modify a few tokens in the seed prompt template to make them slightly token misalignment. 
    \item Misalignment more ($\h_{more}$): modify the token order in the seed prompt template to make them significant token misalignment.
    \item Typographical errors ($\h_{typo}$): apply keyboard-level typos (insertion, omission, transposition, or substitution) to up to $k$ randomly-selected words in the seed prompt.
    \item Orthographic errors ($\h_{orth}$): apply $k$ surface-level formatting perturbations (extra spaces, missing spaces, random case flips, or extra punctuation) to the seed prompt.
    \item Paraphrases ($\h_{para}$): use an LLM~\citep[gpt-5.4;][]{openai2025introducinggpt5} to rewrite the seed prompt by replacing exactly $k$ words with semantically equivalent alternatives.
\end{enumerate}
Types 5 to 7 are applied to the prompt body rather than the template. A word here means an alphabetic token of at least three letters. For $\h_{typo}$, each inserted or substituted character is drawn from the QWERTY-neighbor set of the original character, with letter case preserved. For $\h_{orth}$, each perturbation is drawn uniformly from four operations, duplicating an existing space, removing the space after a sentence-ending period, flipping the case of a single letter inside a word, or inserting one of `,' / `.' / `;' / `:' after a word, none of which inserts, deletes, or substitutes letters within a word. For $\h_{para}$, the LLM returns the rewritten prompt together with $k$ (original, replacement) pairs; outputs that fail the word-count check are retried, and results are cached so that the same paraphrase is reused across all 11 models.

We randomly select 500 samples from each dataset. Types 1 to 4 are evaluated on the four MCQ datasets only, as they are defined on a fixed MCQ template; types 5 to 7 are additionally evaluated on the open-ended generation dataset Alpaca. For modification types 1 to 4, we create three different variants per type. For types 5 to 7, we vary the severity $k \in \{1, 2, 3\}$ and produce one variant per $(type, k)$ pair, yielding 3 variants per type.\footnote{Here, $type$ refers to one of $\h_{typo}$, $\h_{orth}$, or $\h_{para}$; each type thus consists of three variants, one per $k \in \{1, 2, 3\}$.}
We refer to the seed prompt template as $\h_{seed}$ and the seven modified versions as $\h_{first}$, $\h_{latter}$, $\h_{fewer}$, $\h_{more}$, $\h_{typo}$, $\h_{orth}$, and $\h_{para}$, respectively.
From Eq.~(\ref{eq:ineq}), when $\h_0$ is fixed, the upper bound of $|\deltaLogP|$ is determined by $\|\deltaH\|$. 
This implies that a higher $\|\deltaH\|$ imposes a looser constraint on $|\deltaLogP|$. 
Therefore, we calculate $\|\deltaH\|$ between $\h_{seed}$ and the seven types of modifications for comparison.

Figure~\ref{fig:result_of_rq2} shows the comparison results.
Experimental results show that when comparing $\h_{first}$ and $\h_{latter}$, smaller models such as \texttt{Pythia-1B} (Figure~\ref{fig:result_of_rq2_1}) exhibit a higher $\|\deltaH\|$ of $\h_{latter}$ than that of $\h_{first}$. However, as model size increases, such as in \texttt{Llama3.2-3B} (Figure~\ref{fig:result_of_rq2_2}), the $\|\deltaH\|$ of $\h_{latter}$ becomes comparable to that of $\h_{first}$. When model size increases to 4B, such as \texttt{Qwen1.5-4B} (Figure~\ref{fig:result_of_rq2_3}), $\h_{first}$'s $\|\deltaH\|$ surpasses $\h_{latter}$'s $\|\deltaH\|$. 
Within the \texttt{Qwen1.5} and \texttt{Llama3.2} series, smaller models are more sensitive to latter-half modifications, while larger models are more sensitive to first-half modifications. This size-dependent transition is not observed in the older \texttt{GPT2} and \texttt{Pythia} series (Appendix~\ref{sec:more_results_of_rq2}).

Figure~\ref{fig:result_of_rq2_4} shows the comparison results between tokens with fewer ($\h_{fewer}$) and more ($\h_{more}$) misalignments in the prompt. We observe that different numbers of token misalignments produce significant differences in $\deltaH$.
Specifically, $\h_{more}$ is more likely than $\h_{fewer}$ to lead to prompt sensitivity.

In addition, we compare the trends of $\|\deltaH\|$ for modification and misalignment. Here, modification refers to the average result of $\h_{first}$ and $\h_{latter}$, while misalignment denotes the average result of $\h_{fewer}$ and $\h_{more}$. 
As shown in Figure~\ref{fig:result_of_rq2_5}, misaligned prompt tokens are more likely to lead to prompt sensitivity than modified prompt tokens.

Figure~\ref{fig:result_of_rq2_6} compares $\h_{typo}$, $\h_{orth}$, and $\h_{para}$ on the Alpaca dataset. We observe that $\h_{typo}$ induces the highest $\|\deltaH\|$, followed by $\h_{orth}$ and then $\h_{para}$. This indicates that character-level typographical errors are more likely to lead to prompt sensitivity than surface-level orthographic perturbations or word-level paraphrases, as typos disrupt subword tokenization most aggressively while paraphrases preserve the majority of tokens.
See Appendix~\ref{sec:more_results_of_rq2} for more experimental results.

\subsection{What Is the Relationship Between the Upper Bound and an Existing Prompt Sensitivity Metric? (RQ3)}
Many studies~\cite{zhuo-etal-2024-prosa,chatterjee-etal-2024-posix} propose prompt sensitivity metrics to evaluate the prompt sensitivity of LLMs.
These metrics are typically single values that represent the sensitivity of LLMs to different meaning-preserving prompt templates.
According to Eq.~(\ref{eq:ineq}), a higher upper bound makes it more difficult for LLMs to achieve $|\deltaLogP|$ close to 0.
Therefore, what is the relationship between the upper bound and the prompt sensitivity of LLMs? To answer this question, we compare the prompt sensitivity metric \texttt{PromptSensiScore}~\citep[PSS;][]{zhuo-etal-2024-prosa} with the upper bound.
PSS is a value ranging from 0 to 1, where a lower value indicates lower prompt sensitivity.\footnote{The calculation method for PSS is provided in Appendix~\ref{sec:pss}.} Theoretically, LLMs with smaller upper bounds should have a higher chance of achieving lower PSS.
Figure~\ref{fig:result_of_rq3} shows the relationship between the average upper bound across layers of LLMs and PSS.
We can observe that within the same model series, the average upper bound is positively correlated with PSS.
This indicates that the smaller the average upper bound, the greater the chance that LLMs achieve lower prompt sensitivity.
The positive correlation between upper bounds and PSS further indicates that upper bounds influence the prompt sensitivity of LLMs.

\subsection{Which Factor Contributes to the Change of Logits? (RQ4)}

\begin{figure}[t]
  \includegraphics[width=\columnwidth]{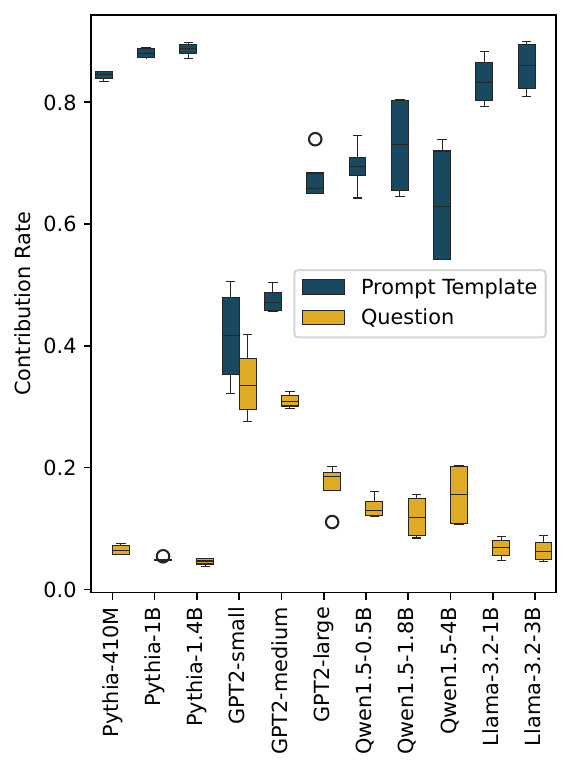}
  \caption{Comparison of contribution rates of prompt templates and questions to logits.}
  \label{fig:result_of_rq4}
\end{figure}

\citet{wu-varshney-2025-transformer}'s study indicates that LLMs tend to cluster the same task samples. This inspires us to evaluate whether the outputs of LLMs are more influenced by the prompt template or the question itself. 
In particular, we construct an ordinary least squares regression model that uses different prompt templates and questions to predict the logit of the model's next token.
Subsequently, we perform an analysis of variance (ANOVA) on the regression results to calculate each factor's contribution to the total variance and further determine each factor's proportion of contribution relative to the total sum of squares.
Figure~\ref{fig:result_of_rq4} compares the contributions of the prompt template and question to the logits.
The prompt template is the primary factor explaining the logit variation.
Except for \texttt{GPT2-small} and \texttt{GPT2-medium}, the contribution rate of prompt templates to logits significantly exceeds that of questions.
In addition, the \texttt{Pythia} series model exhibits the lowest variance in contribution rate across all datasets, while the \texttt{Qwen1.5} series model shows the highest variance in contribution rate across all datasets. This indicates that the \texttt{Qwen1.5} series model is significantly sensitive to different datasets.

\section{Related Work}

\subsection{Prompt Sensitivity of LLMs}
LLMs have strong in-context learning capabilities~\citep{NEURIPS2020_1457c0d6}, enabling them to perform diverse tasks based on prompts, often without requiring additional fine-tuning~\citep{Radford2019LanguageMA,JMLR:v21:20-074,gao-etal-2021-making}.
However, the stability and reliability of this learning approach remain controversial~\citep{weber2023icl}. Existing studies indicate that model outputs are highly dependent on multiple factors, such as the choice and order of examples~\citep{liu-etal-2022-makes,su2023selective,lu-etal-2022-fantastically,pmlr-v139-zhao21c}, the definition of input labels~\citep{min-etal-2022-rethinking}, and the phrasing of prompts~\citep{gu-etal-2023-robustness,sun2024evaluating}.
Beyond these factors, LLMs exhibit extreme sensitivity to minor changes in prompt structure or phrasing, even when such alterations preserve semantic meaning. 
This phenomenon has been systematically explored in numerous studies~\citep{voronov-etal-2024-mind,mizrahi-etal-2024-state}, indicating that subtle modifications to prompts can significantly impact model outputs. 
Furthermore, to characterize and compare the prompt sensitivity of different models, numerous studies~\citep{zhuo-etal-2024-prosa,chatterjee-etal-2024-posix} have constructed specialized benchmarks to quantify and evaluate models' robustness to prompt perturbations. 
Contrary to previous work, this study attempts to represent LLMs as functions, leveraging Taylor expansion to explain the mechanism behind prompt sensitivity from the function perspective.
It provides both theoretical foundations and empirical evidence to explain why LLMs exhibit prompt sensitivity.

\subsection{LLMs as Functions}
In recent years, some studies have attempted to characterize LLMs from the perspective of function mapping~\citep{NEURIPS2020_1457c0d6,wei2022chain}.
This perspective abstracts an LLM as a function mapping $x$ to a distribution over $y$. In other words, given a prompt $x$, the model defines a distribution $P(y\mid x)$ for an output $y$.
This functional representation facilitates a unified understanding of model behavior across different tasks and provides a theoretical framework for analyzing LLM generalization and robustness.
Notably, it has also been shown that transformers themselves serve as universal approximators of sequence-to-sequence functions~\citep{Yun2020Are}, further reinforcing the perspective that LLMs are functions.
Building on this idea of function mapping, some studies consider prompt engineering as a design problem for function call interfaces, investigating how different prompt formats alter the properties of the function mapping~\citep{liu2023pre}. In our study, we consider LLMs as composite functions that can be split into the feature processing part and the function part between any transformer blocks. This split allows us to perform a Taylor expansion on any part of the models for analysis.

\section{Conclusion}
Prompt sensitivity, which describes how LLMs produce different outputs in response to meaning-preserving prompts, raises user concerns about the stability and reliability of LLMs.
To investigate the underlying mechanisms of prompt sensitivity and to better understand LLMs, we started by considering LLMs as multivariate continuous functions. We pointed out that improving classification accuracy requires internal clustering behavior within neural networks. Then, we applied the first-order Taylor expansion to LLMs. 
By observing changes in hidden states across all layers, we found that transformer-based LLMs lacked this clustering behavior, leading to a high upper bound on the difference in log probabilities between two prompts. We also identified which types of modifications are more likely to lead to prompt sensitivity. Moreover, the upper bound of the difference in log probabilities correlated positively with an existing prompt sensitivity metric. Counterintuitively, our analysis revealed that prompt templates contributed more significantly to logits than the questions themselves.
In the future, we will attempt to introduce higher-order Taylor terms (such as second-order terms implemented via the Hessian matrix) to achieve more precise and faithful bounds.
We also plan to extend this work to analyze the entire log probability space and multi-step generation process.

\section*{Limitations}
One limitation of this work is we only considered the log probabilities of a single dimension for the model's next token, implicitly requiring the entire logit distribution of the next token to remain consistent across meaning-preserving prompts.
This requirement poses a significant challenge for LLMs.
Moreover, we employed only a first-order Taylor expansion.
Given that LLMs are naturally highly complex functions, this linear approximation may introduce some errors. 
In the future, exploring higher-order Taylor expansions could yield more precise approximations.

\section*{Acknowledgments}
This work was supported by JST BOOST, Grant Number JPMJBS2407.
We thank the constructive comments from the anonymous reviewers, which helped improve this work.
We also appreciate the careful attention of the meta reviewer.

\bibliography{custom}

\appendix

\newpage
\section{Taylor Expansion}
\label{sec:taylor_expansion}

\subsection{Background} 
The roots of Taylor expansion can be traced back to early thoughts on infinity, such as the paradoxes of divisibility proposed by the ancient Greek philosopher Zeno~\citep{Lindberg1992}, as well as the ``method of exhaustion'' developed by Archimedes~\citep{heath1981history} and later by Liu Hui~\citep{martzloff2007history}, which laid the foundation for approximating infinite processes through finite steps.
In the 14th century, Indian mathematician Madhava of Sangamagrama and his successors in the Kerala school developed series expansions for functions such as sine, cosine, and arctangent, marking the earliest concrete examples of power series methods analogous to later Taylor expansions~\citep{Lindberg1992}. 
In the 17th century, Newton and Gregory independently developed general methods for expanding functions~\citep{inglis1940james}. 
Later, Brook Taylor first systematically proposed an expansion method applicable to general functions in 1715, forming the basis of today's Taylor expansions~\citep{taylor1715methodus}.
In our study, we consider LLMs as functions and employ first-order Taylor expansions to connect prompts, their gradients, and the logit of the model's next token, thereby analyzing the constraint relationships among them.

\subsection{The First-order Taylor Expansion}
In mathematics, the Taylor series or Taylor expansion of a function is an infinite sum of terms that are expressed in terms of the function's derivatives at a single point. The partial sum formed by the first $n + 1$ terms of a Taylor series is a polynomial of degree n that is called the nth Taylor polynomial of the function. Taylor polynomials are approximations of a function, which become generally more accurate as $n$ increases. The first-order Taylor expansion in one variable of $f(x)$ about $x=a$  is as follows:
\begin{equation}
\begin{aligned}
    f(x) = &f(a)+f'(a)(x-a)\\ &+\mathcal{O}((x-a)^2)\quad (x\to a).
\end{aligned}
\end{equation}
where $\mathcal{O}(x-a)$ indicates the infinitesimal term of higher order than $(x-a)$, and $x \to a$ indicates that this equality holds as $x$ approaches $a$. In other words, this expansion is a local approximation describing the behavior of $f(x)$ near $x = a$.

For more complex multivariate scenarios, we suppose $f:\mathbb{R}^n \to \mathbb{R}$ is differentiable at the point $\boldsymbol{a}=(a_1,a_2,\dots,a_n)$. Then the first-order Taylor expansion of f at $\boldsymbol{x}=(x_1,x_2,\dots,x_n)$ is:
\begin{equation}
\begin{aligned}
    f(\boldsymbol{x}) &= f(\boldsymbol{a}) + \nabla f(\boldsymbol{a}) \cdot (\boldsymbol{x}-\boldsymbol{a}) \\ &+ \mathcal{O}(\|\boldsymbol{x}-\boldsymbol{a}\|^2) \quad (\boldsymbol{x} \to \boldsymbol{a}).
\end{aligned}
\end{equation}
where $\nabla f(\boldsymbol{a}) = \left( \frac{\partial f}{\partial x_1}(\boldsymbol{a}), \frac{\partial f}{\partial x_2}(\boldsymbol{a}), \dots, \frac{\partial f}{\partial x_n}(\boldsymbol{a}) \right)$ is the gradient.
In the expression $\mathcal{O}(\|\boldsymbol{x}-\boldsymbol{a}\|)$, the norm $\|\cdot\|$ can be any norm (such as the Euclidean norm (2-norm) or vector norm) on $\mathbb{R}^n$, because all norms in finite-dimensional spaces are equivalent. The $\mathcal{O}(\|\boldsymbol{x}-\boldsymbol{a}\|^2)$ means the remainder term that vanishes faster than $\|\boldsymbol{x}-\boldsymbol{a}\|^2$ as $\boldsymbol{x} \to \boldsymbol{a}$. The operator $\cdot$ denotes the dot product.

\section{Proof}
\label{sec:proof}

\textbf{Statement}. Suppose $\vx_i$ and $\vx_j$ are normalized unit vectors, i.e., $\|\vx_i\|^2=\|\vx_j\|^2=1$, and $\theta_{ij}$ is the angle between them, then the following holds:
\begin{equation}
\label{eq:app5}
    \|\vx_i-\vx_j\| = \sqrt{2-2\cos \theta_{ij}}
\end{equation}

\textit{Proof.} As $\vx_i$ and $\vx_j$ are unit vectors, 
\begin{align}
    \|\vx_i-\vx_j\|^2
    &= \langle \vx_i-\vx_j,\ \vx_i-\vx_j \rangle \\
    &= \|\vx_i\|^2+\|\vx_j\|^2-2\langle \vx_i,\vx_j\rangle \\
    &= 1+1-2\langle \vx_i,\vx_j\rangle \\
    &= 2-2\|\vx_i\|\,\|\vx_j\|\cos\theta_{ij} \\
    &= 2-2\cos\theta_{ij}.
\end{align}

Taking square roots yields: $\|\vx_i-\vx_j\| = \sqrt{2-2\cos \theta_{ij}}$.


\section{Hyperparameters for Training ResNet.}
\label{sec:the_hyperparameters_for_training_resnet}

To ensure stable optimization and efficient convergence of the \texttt{ResNet-101} network on the CIFAR-10 dataset, a carefully designed hyperparameter configuration scheme was employed during training.

\begin{figure*}[htbp]
    \centering
    \includegraphics[width=\textwidth]{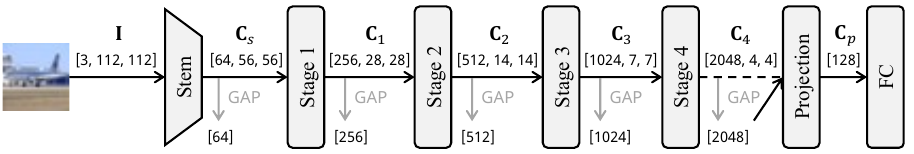}
    \caption{The feature maps' shape of the neural network. The batch size dimension is omitted and ``\textcolor{gray}{GAP}'' indicates global mean pooling.}
    \label{fig:resnet_shape}
\end{figure*}

As shown in Figure~\ref{fig:resnet_shape}, our network architecture is a ResNet connected to a projection layer and a fully connected layer. This section provides more details.
We preprocess images using the following pipeline before feeding them into ResNet:
\begin{verbatim}
transform = transforms.Compose([
    transforms.Resize(112), 
    transforms.CenterCrop(112),
    transforms.ToTensor()
])
\end{verbatim}
We project the 2048-dimensional features from ResNet's stage 4 output onto a 128-dimensional embedding space, then classify them using a fully connected (classification) layer. The specific network architecture is as follows:
\begin{verbatim}
proj = nn.Sequential(
    nn.Linear(2048, 512),
    nn.BatchNorm1d(512),
    nn.PReLU(),
    nn.Dropout(p=0.2),
    nn.Linear(512, 128)
)

clf = nn.Linear(128, num_classes)
\end{verbatim}
Our experiment employs the cross-entropy loss function with the AdamW optimizer~\citep{loshchilov2017decoupled}, using the macro F1 score as the primary evaluation metric. The training process utilizes a batch size of 128 and runs for 20 epochs.

\paragraph{Input and Output Shapes.} As shown in Figure~\ref{fig:resnet_shape}, we mark the shape of each feature map. Here, ``\textcolor{gray}{GAP}'' denotes the global average pooling operation. After performing $L^2$ normalization on the vector obtained from global average pooling, the distance is calculated using Eq.~(\ref{eq:5}). It is especially worth noting that the output $\tC_4$ of ``Stage 4'' undergoes global average pooling before being fed into the ``Projection'' layer as input.

\section{Prompt Sensitivity Metric: PSS}
\label{sec:pss}
In this section, we introduce PSS~\cite{zhuo-etal-2024-prosa}. For each set of all prompt variants under the same question, we have:
\begin{equation}
   S = \frac{\sum_{p_i,p_j \in P} (|Y(P_i)-Y(P_j)|)}{C(|P|,2)}, 
\end{equation}
where $Y(p)$ represents the performance metric under prompt $p$. In our study, $Y(p)$ refers to correctness.
$|Y(P_i)-Y(P_j)|$ represents the absolute value difference in performance metrics between prompt $p_i$ and prompt $p_j$. $C(|P|,2)$ represents the count of prompt pairs in the same question. PSS is given by the following:
\begin{equation}
    PSS = \frac{1}{N}\sum_{i=1}^{N} S_i,
\end{equation}
where $N$ is the total number of questions and $S_i$ is the score for the $i$-th question.

\section{Prompt Templates}
\label{sec:prompt_templates}
\subsection{Meaning-Preserving Prompt Templates}
\label{sec:meaning-preserving_prompts}
In this section, we provide the 12 prompt templates provided by~\citet{zhuo-etal-2024-prosa} mentioned in \S~\ref{sec:experimental_verifications}. For multiple-choice questions with 4 options, the templates are shown in Table~\ref{tab:12-templates}.
We choose 12 prompts for experimentation to ensure data diversity and avoid inaccurate results caused by individual edge cases.

\begin{table*}[t]
\caption{The meaning-preserving prompt templates for ARC Challenge, MMLU, and OpenBookQA datasets. For the CommonSenseQA dataset, the number of options changes from four to five, so option `E' should be added accordingly.
\textcolor{gray}{Gray} text indicates template slots that need to be replaced.}
\label{tab:12-templates}
\begin{center}
\begin{tabular}{lp{13cm}}
\toprule
\texttt{prompt 1}         &  \texttt{\textcolor{gray}{\textcolor{gray}{\{question\}}}\cn A. \textcolor{gray}{\{A\}}\cn B. \textcolor{gray}{\{B\}}\cn C. \textcolor{gray}{\{C\}}\cn D. \textcolor{gray}{\{D\}}\cn Answer:} \\
\\
\texttt{prompt 2}         & \texttt{Question:\cn \textcolor{gray}{\{question\}}\cn A. \textcolor{gray}{\{A\}}\cn B. \textcolor{gray}{\{B\}}\cn C. \textcolor{gray}{\{C\}}\cn D. \textcolor{gray}{\{D\}}\cn Answer:} \\
\\
\texttt{prompt 3}         & \texttt{Question:\cn \textcolor{gray}{\{question\}} A. \textcolor{gray}{\{A\}} B. \textcolor{gray}{\{B\}} C. \textcolor{gray}{\{C\}} D. \textcolor{gray}{\{D\}}\cn Answer:} \\
\\
\texttt{prompt 4}         & \texttt{Could you provide a response to the following question: \textcolor{gray}{\{question\}} A. \textcolor{gray}{\{A\}} B. \textcolor{gray}{\{B\}} C. \textcolor{gray}{\{C\}} D. \textcolor{gray}{\{D\}}} \\
\\
\texttt{prompt 5}         & \texttt{Please answer the following question:\cn\textcolor{gray}{\{question\}}\cn A. \textcolor{gray}{\{A\}}\cn B. \textcolor{gray}{\{B\}}\cn C. \textcolor{gray}{\{C\}}\cn D. \textcolor{gray}{\{D\}}} \\
\\
\texttt{prompt 6}         & \texttt{Please address the following question:\cn \textcolor{gray}{\{question\}}\cn A. \textcolor{gray}{\{A\}}\cn B. \textcolor{gray}{\{B\}}\cn C. \textcolor{gray}{\{C\}}\cn D. \textcolor{gray}{\{D\}}\cn Answer:} \\
\\
\texttt{prompt 7}         & \texttt{You are a very helpful AI assistant. Please answer the following questions: \textcolor{gray}{\{question\}} A. \textcolor{gray}{\{A\}} B. \textcolor{gray}{\{B\}} C. \textcolor{gray}{\{C\}} D. \textcolor{gray}{\{D\}}} \\
\\
\texttt{prompt 8}         & \texttt{As an exceptionally resourceful AI assistant, I'm at your service. Address the questions below:\cn \textcolor{gray}{\{question\}}\cn A. \textcolor{gray}{\{A\}}\cn B. \textcolor{gray}{\{B\}}\cn C. \textcolor{gray}{\{C\}}\cn D. \textcolor{gray}{\{D\}}} \\
\\
\texttt{prompt 9}         & \texttt{As a helpful Artificial Intelligence Assistant, please answer the following questions\cn \textcolor{gray}{\{question\}} A. \textcolor{gray}{\{A\}}\cn B. \textcolor{gray}{\{B\}}\cn C. \textcolor{gray}{\{C\}}\cn D. \textcolor{gray}{\{D\}}} \\
\\
\texttt{prompt 10}         & \texttt{Could you provide a response to the following question: \textcolor{gray}{\{question\}} A. \textcolor{gray}{\{A\}} B. \textcolor{gray}{\{B\}} C. \textcolor{gray}{\{C\}} D. \textcolor{gray}{\{D\}}\cn Answer the question by replying A, B, C or D.} \\
\\
\texttt{prompt 11}         & \texttt{Please answer the following question:\cn\textcolor{gray}{\{question\}}\cn A. \textcolor{gray}{\{A\}}\cn B. \textcolor{gray}{\{B\}}\cn C. \textcolor{gray}{\{C\}}\cn D. \textcolor{gray}{\{D\}}\cn Answer the question by replying A, B, C or D.} \\
\\
\texttt{prompt 12}         & \texttt{Please address the following question:\cn\textcolor{gray}{\{question\}}\cn A. \textcolor{gray}{\{A\}}\cn B. \textcolor{gray}{\{B\}}\cn C. \textcolor{gray}{\{C\}}\cn D. \textcolor{gray}{\{D\}}\cn Answer this question by replying A, B, C or D.} \\
\bottomrule
\end{tabular}
\end{center}
\end{table*}

\subsection{Modification and Misalignment Prompt Templates}
\label{sec:modification_and_misalignment_prompt_templates}

To evaluate which types of prompts may lead to higher prompt sensitivity, we create four prompt templates for quantitative analysis. These four prompt templates are shown in Table~\ref{tab:4-templates}.
These prompt templates are modified from a seed prompt template, which is: ``\texttt{You are a very helpful AI assistant. Please answer the following questions:\cn Question: \textcolor{gray}{\{question\}}\cn A. \textcolor{gray}{\{A\}} B. \textcolor{gray}{\{B\}} C. \textcolor{gray}{\{C\}} D. \textcolor{gray}{\{D\}}\cn Please choose the best option and respond only with the option of the correct answer (A, B, C, or D).\cn Answer:}''

Our experimental implementation process is as follows:  We first randomly select 500 samples from each of the four datasets.  We then combine these samples with both the seed prompt template and our modified 12 prompt templates, creating 6,500 prompts for each dataset.  These prompts feed into the LLMs for testing.

\begin{table*}[h]
\caption{Our prompt templates for ARC Challenge, MMLU, and OpenBookQA datasets. For the CommonSenseQA dataset, the number of options changes from four to five, so option `E' should be added accordingly.
\textcolor{gray}{Gray} text indicates template slots that need to be replaced. 
\textcolor{green}{Green} indicates the modified token in the first half of the prompt.
\textcolor{red}{Red} indicates the modified token in the latter half of the prompt.
\textcolor{orange}{Orange} indicates the token causing the misalignment in the prompt.
\textcolor{blue}{Blue} indicates that the prompt is completely misaligned.
}
\label{tab:4-templates}
\begin{center}
\scalebox{0.75}{
\begin{tabular}{lp{13cm}}
\toprule
\texttt{Seed prompt}         &  \texttt{You are a very helpful AI assistant. Please answer the following questions:\cn Question: \textcolor{gray}{\{question\}}\cn A. \textcolor{gray}{\{A\}} B. \textcolor{gray}{\{B\}} C. \textcolor{gray}{\{C\}} D. \textcolor{gray}{\{D\}}\cn Please choose the best option and respond only with the option of the correct answer (A, B, C, or D).\cn Answer:
}\\
\hline

\multirow{14}{*}{\texttt{Modification first}} & \texttt{You are a very \textcolor{green}{useful} AI assistant. Please answer the following questions:\cn Question: \textcolor{gray}{\{question\}}\cn A. \textcolor{gray}{\{A\}} B. \textcolor{gray}{\{B\}} C. \textcolor{gray}{\{C\}} D. \textcolor{gray}{\{D\}}\cn Please choose the best option and respond only with the option of the correct answer (A, B, C, or D).\cn Answer:
} \\
\\
& \texttt{You are a very \textcolor{green}{smart} AI assistant. Please answer the following questions:\cn Question: \textcolor{gray}{\{question\}}\cn A. \textcolor{gray}{\{A\}} B. \textcolor{gray}{\{B\}} C. \textcolor{gray}{\{C\}} D. \textcolor{gray}{\{D\}}\cn Please choose the best option and respond only with the option of the correct answer (A, B, C, or D).\cn Answer:
} \\
\\
& \texttt{You are a very \textcolor{green}{friendly} AI assistant. Please answer the following questions:\cn Question: \textcolor{gray}{\{question\}}\cn A. \textcolor{gray}{\{A\}} B. \textcolor{gray}{\{B\}} C. \textcolor{gray}{\{C\}} D. \textcolor{gray}{\{D\}}\cn Please choose the best option and respond only with the option of the correct answer (A, B, C, or D).\cn Answer:
} \\
\hline

\multirow{14}{*}{\texttt{Modification latter}} & \texttt{You are a very helpful AI assistant. Please answer the following questions:\cn Question: \textcolor{gray}{\{question\}}\cn A. \textcolor{gray}{\{A\}} B. \textcolor{gray}{\{B\}} C. \textcolor{gray}{\{C\}} D. \textcolor{gray}{\{D\}}\cn Please choose the best option and respond only with the option of the \textcolor{red}{suitable} answer (A, B, C, or D).\cn Answer:
} \\
\\
& \texttt{You are a very helpful AI assistant. Please answer the following questions:\cn Question: \textcolor{gray}{\{question\}}\cn A. \textcolor{gray}{\{A\}} B. \textcolor{gray}{\{B\}} C. \textcolor{gray}{\{C\}} D. \textcolor{gray}{\{D\}}\cn Please choose the best option and respond only with the \textcolor{red}{letter} of the correct answer (A, B, C, or D).\cn Answer:
} \\
\\
& \texttt{You are a very helpful AI assistant. Please answer the following questions:\cn Question: \textcolor{gray}{\{question\}}\cn A. \textcolor{gray}{\{A\}} B. \textcolor{gray}{\{B\}} C. \textcolor{gray}{\{C\}} D. \textcolor{gray}{\{D\}}\cn Please choose the best option and respond only with the \textcolor{red}{choice} of the correct answer (A, B, C, or D).\cn Answer:
} \\
\hline

\multirow{14}{*}{\texttt{Misalignment fewer}} & \texttt{You are a very helpful AI assistant. Please answer the following questions:\cn Question: \textcolor{gray}{\{question\}}\cn A. \textcolor{gray}{\{A\}} B. \textcolor{gray}{\{B\}} C. \textcolor{gray}{\{C\}} D. \textcolor{gray}{\{D\}}\cn Please choose the best option and respond only with the option of the answer (A, B, C, or D) \textcolor{orange}{below}.\cn Answer:
} \\
\\
& \texttt{You are a very helpful AI assistant. Please answer the following questions:\cn Question: \textcolor{gray}{\{question\}}\cn A. \textcolor{gray}{\{A\}} B. \textcolor{gray}{\{B\}} C. \textcolor{gray}{\{C\}} D. \textcolor{gray}{\{D\}}\cn Please choose the best option and respond only with the option of the answer (A, B, C, or D) \textcolor{orange}{carefully}.\cn Answer:
} \\
\\
& \texttt{You are a very helpful AI assistant. Please answer the following questions:\cn Question: \textcolor{gray}{\{question\}}\cn A. \textcolor{gray}{\{A\}} B. \textcolor{gray}{\{B\}} C. \textcolor{gray}{\{C\}} D. \textcolor{gray}{\{D\}}\cn Please choose the best option and respond only with the option of the answer (A, B, C, or D) \textcolor{orange}{now}.\cn Answer:
}  \\
\hline

\multirow{14}{*}{\texttt{Misalignment more}} & \texttt{\textcolor{blue}{Please choose the best option and respond only with the option of the answer (A, B, C, or D) below.\cn You are a very helpful AI assistant. Please answer the following questions:\cn Question: \textcolor{gray}{\{question\}}\cn A. \textcolor{gray}{\{A\}} B. \textcolor{gray}{\{B\}} C. \textcolor{gray}{\{C\}} D. \textcolor{gray}{\{D\}}\cn Answer:}
} \\
\\
& \texttt{\textcolor{blue}{Please choose the best option and respond only with the option of the answer (A, B, C, or D) carefully.\cn You are a very helpful AI assistant. Please answer the following questions:\cn Question: \textcolor{gray}{\{question\}}\cn A. \textcolor{gray}{\{A\}} B. \textcolor{gray}{\{B\}} C. \textcolor{gray}{\{C\}} D. \textcolor{gray}{\{D\}}\cn Answer:}
} \\
\\
& \texttt{\textcolor{blue}{Please choose the best option and respond only with the option of the answer (A, B, C, or D) now.\cn You are a very helpful AI assistant. Please answer the following questions:\cn Question: \textcolor{gray}{\{question\}}\cn A. \textcolor{gray}{\{A\}} B. \textcolor{gray}{\{B\}} C. \textcolor{gray}{\{C\}} D. \textcolor{gray}{\{D\}}\cn Answer:}
} \\
\bottomrule
\end{tabular}
}
\end{center}
\end{table*}

\section{More Experimental Results}
\label{sec:more_about_experiments}
\subsection{More Results of RQ1}
\label{sec:more_results_of_rq1}

\begin{figure*}[t]
\centering
  \includegraphics[width=0.75\linewidth]{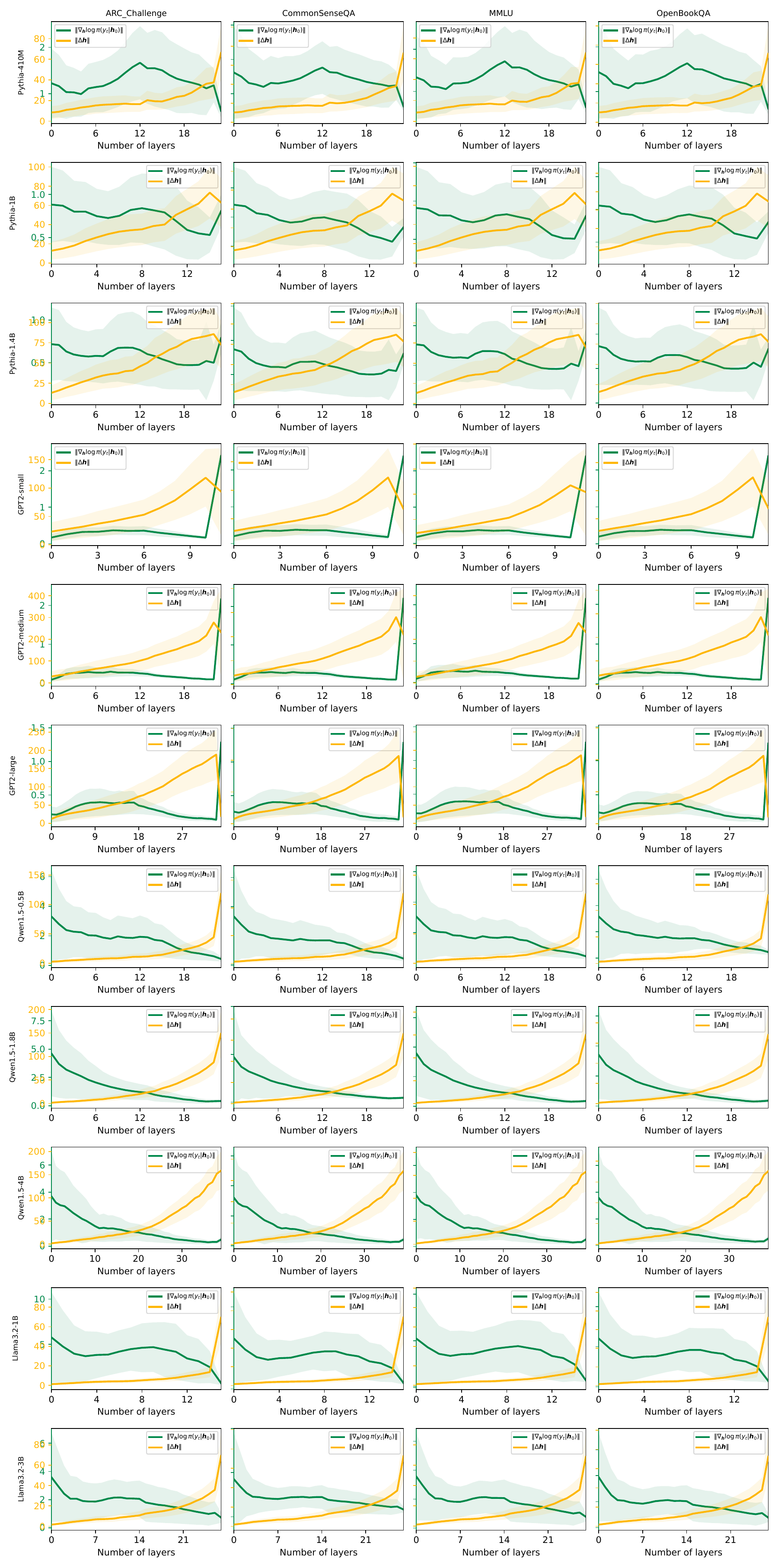}
  \caption{The trends of $\|\gradient\|$ and $\|\deltaH\|$ across layers for all models on the four datasets.}
  \label{fig:full_result_of_rq1_grad_delta_z}
\end{figure*}

\begin{figure*}[t]
\centering
  \includegraphics[width=0.75\linewidth]{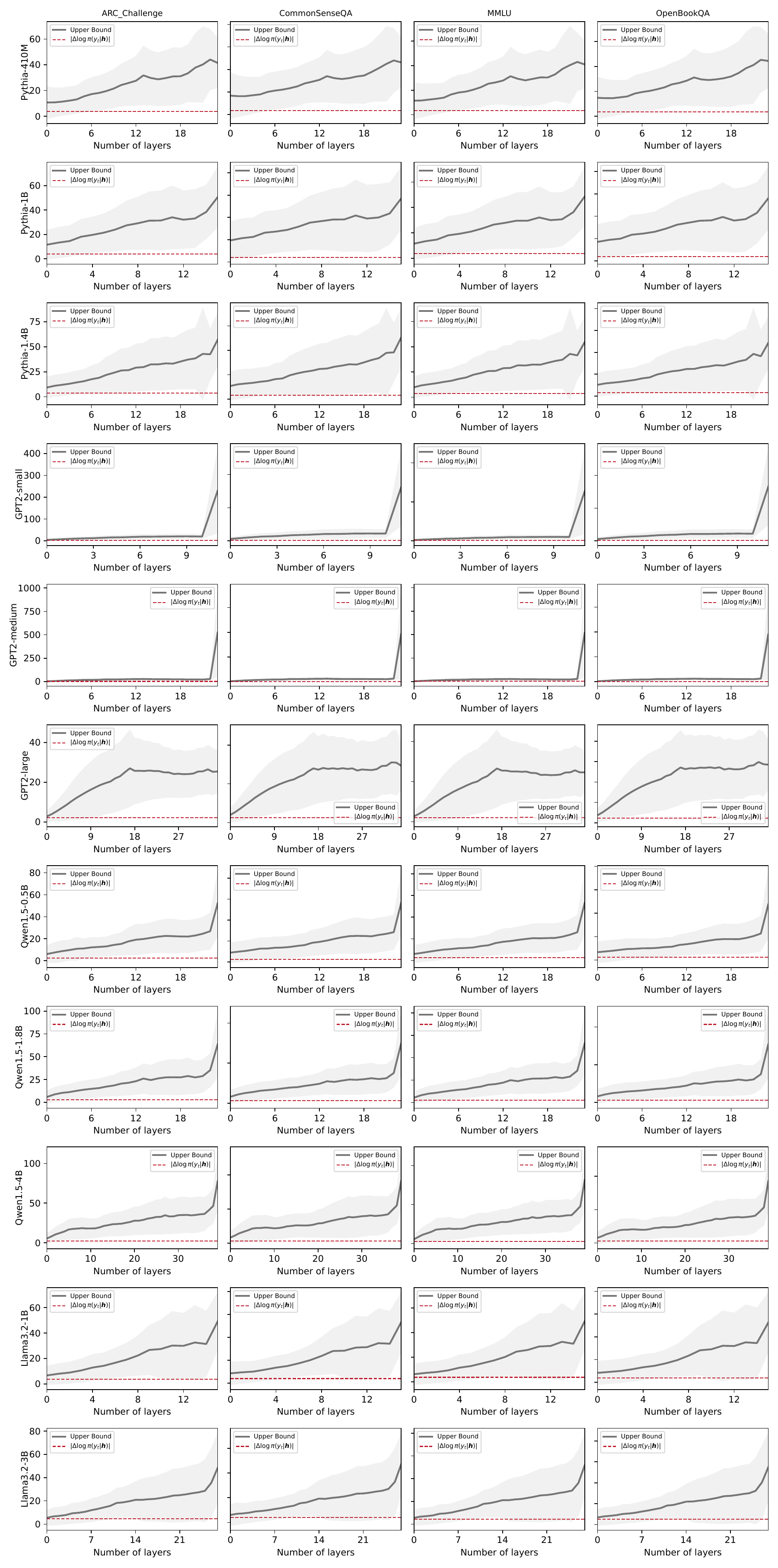}
  \caption{The trends of upper bounds across layers for all models on the four datasets.}
  \label{fig:full_result_of_rq1_upper_bound}
\end{figure*}

Figure~\ref{fig:full_result_of_rq1_grad_delta_z} shows the trends of $\|\gradient\|$ and $\|\deltaH\|$ across layers for all models on the four datasets. 
We can observe that the trends in $\|\gradient\|$ and $\|\deltaH\|$ across models within the same series are similar across all datasets, indicating that our findings are broadly applicable.
Although $\|\deltaH\|$ may suddenly decrease in certain layers of some models (for example, \texttt{Pythia-1B/1.4B} and \texttt{GPT2-small/medium/large}), $\|\gradient\|$ simultaneously increases abruptly to prevent the upper bound from dropping too low.
Figure~\ref{fig:full_result_of_rq1_upper_bound} shows the trends of upper bounds.
Experimental results indicate that upper bounds exhibit increasing trends across all models and datasets, aligning with the conclusion of our RQ1: although gradients decrease across layers, the increase in $\|\deltaH\|$ prevents the upper bound from becoming sufficiently low for $|\deltaLogP|$ to approach 0.

\subsection{More Results of RQ2}
\label{sec:more_results_of_rq2}

\begin{figure*}[t]
\centering
  \includegraphics[width=0.75\linewidth]{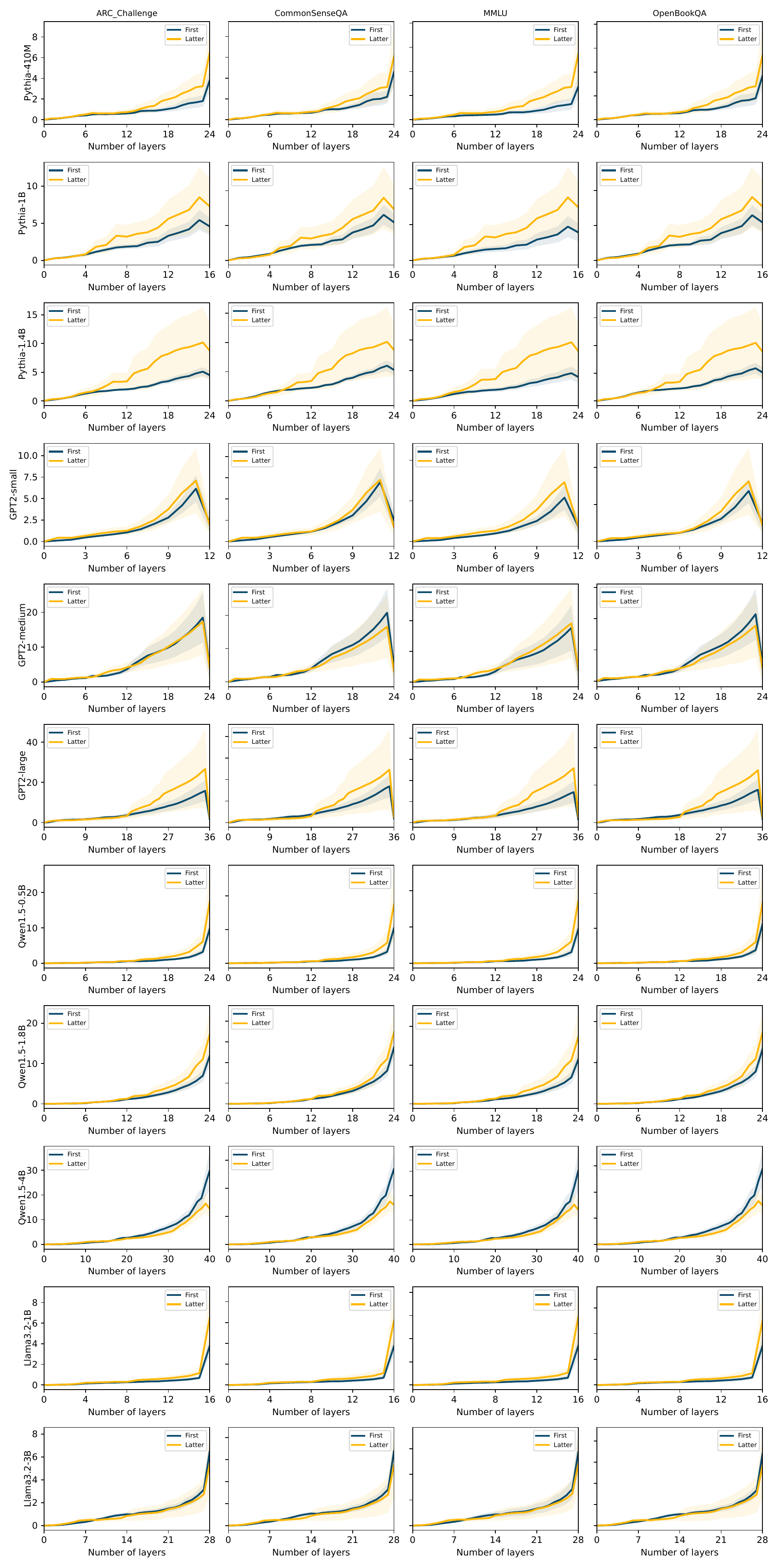}
  \caption{The comparison results between ``First'' and ``Latter'' across all models and datasets.}
  \label{fig:full_result_of_rq2_first_latter}
\end{figure*}

\begin{figure*}[t]
\centering
  \includegraphics[width=0.75\linewidth]{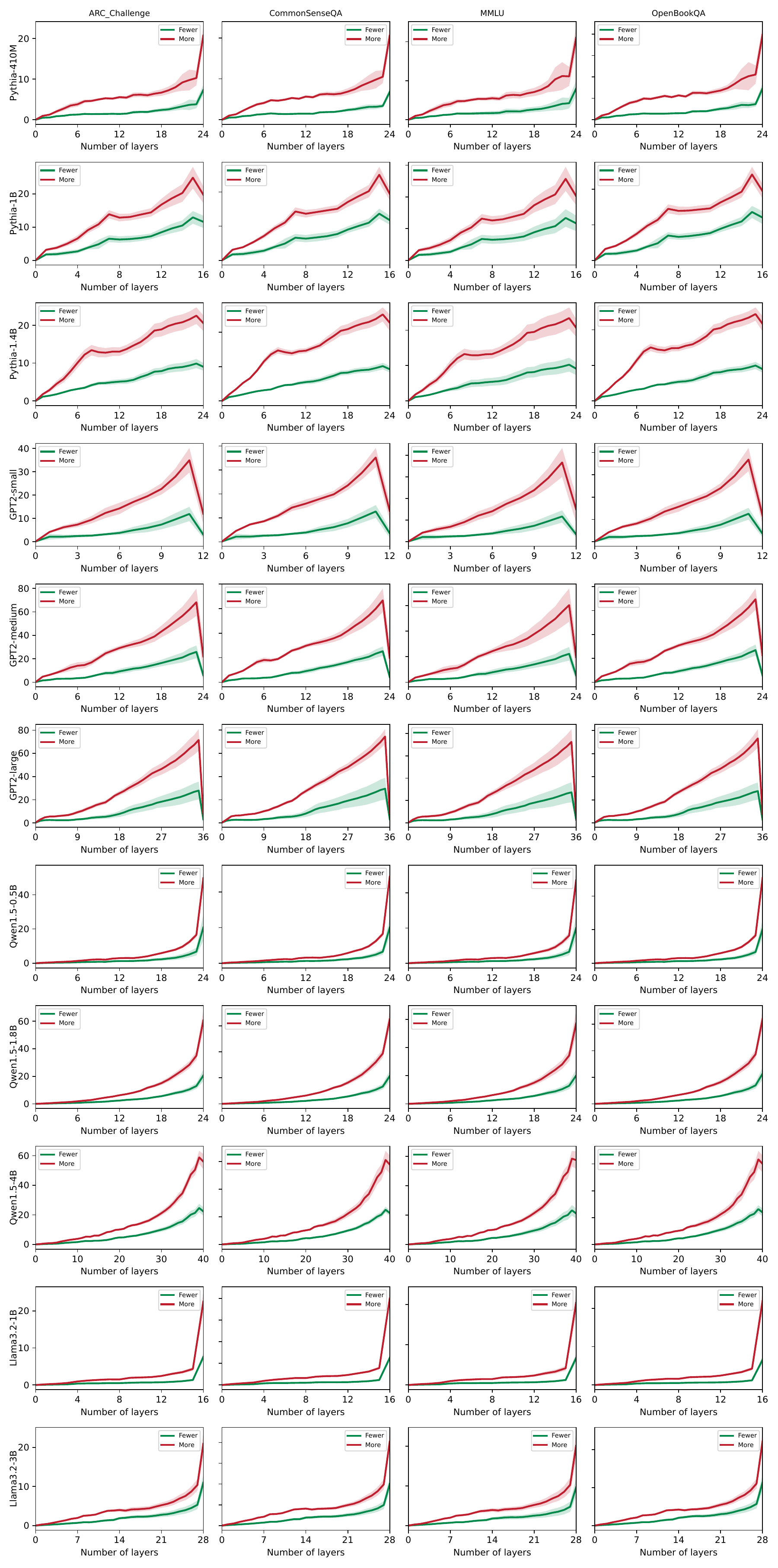}
  \caption{The comparison results between ``Fewer'' and ``More'' across all models and datasets.}
  \label{fig:full_result_of_rq2_fewer_more}
\end{figure*}

\begin{figure*}[t]
\centering
  \includegraphics[width=0.75\linewidth]{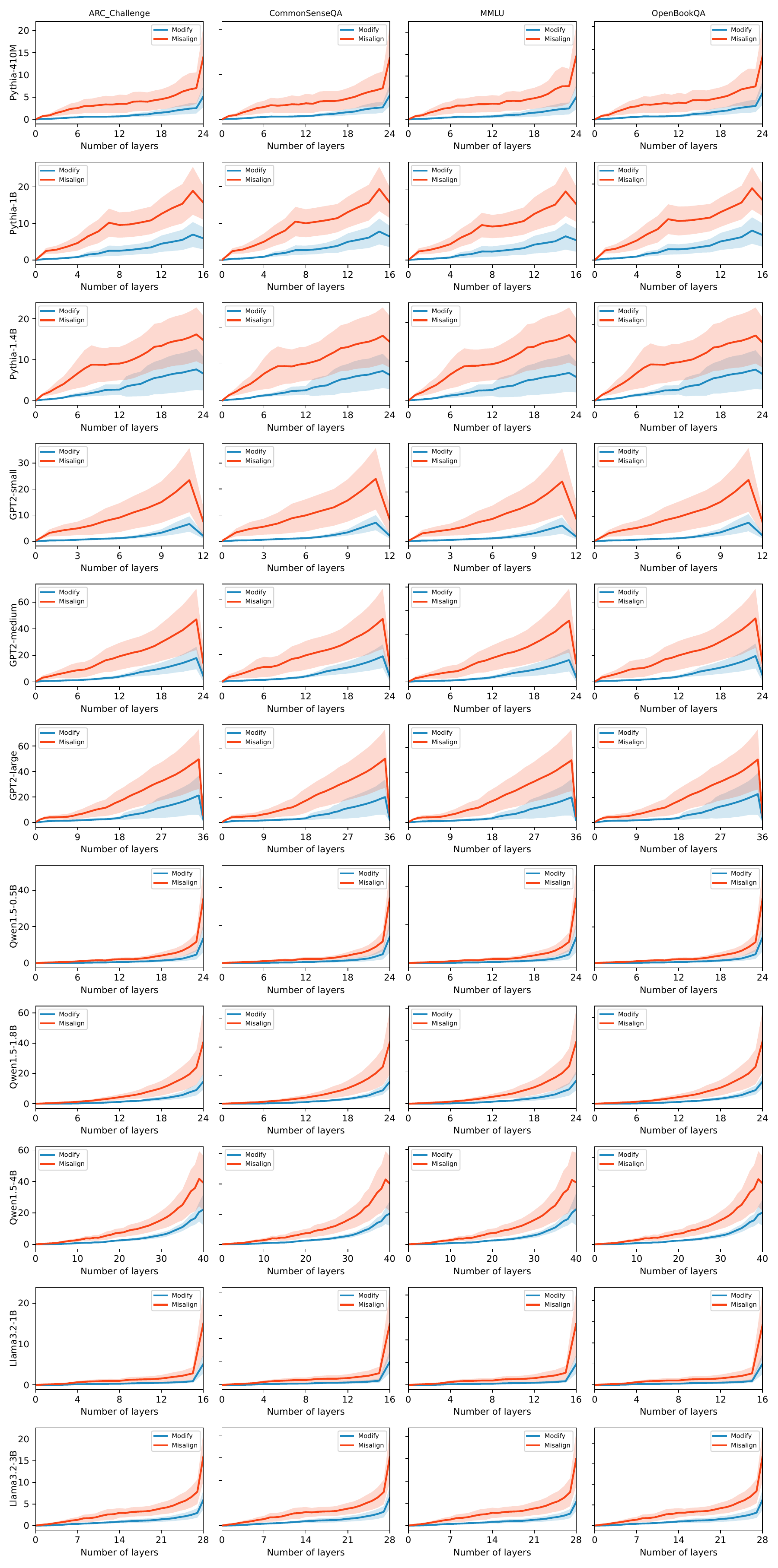}
  \caption{The comparison results between ``Modify'' and ``Misalign'' across all models and datasets.}
  \label{fig:full_result_of_rq2_modify_misalign}
\end{figure*}

\begin{figure*}[t]
\centering
  \includegraphics[width=0.90\linewidth]{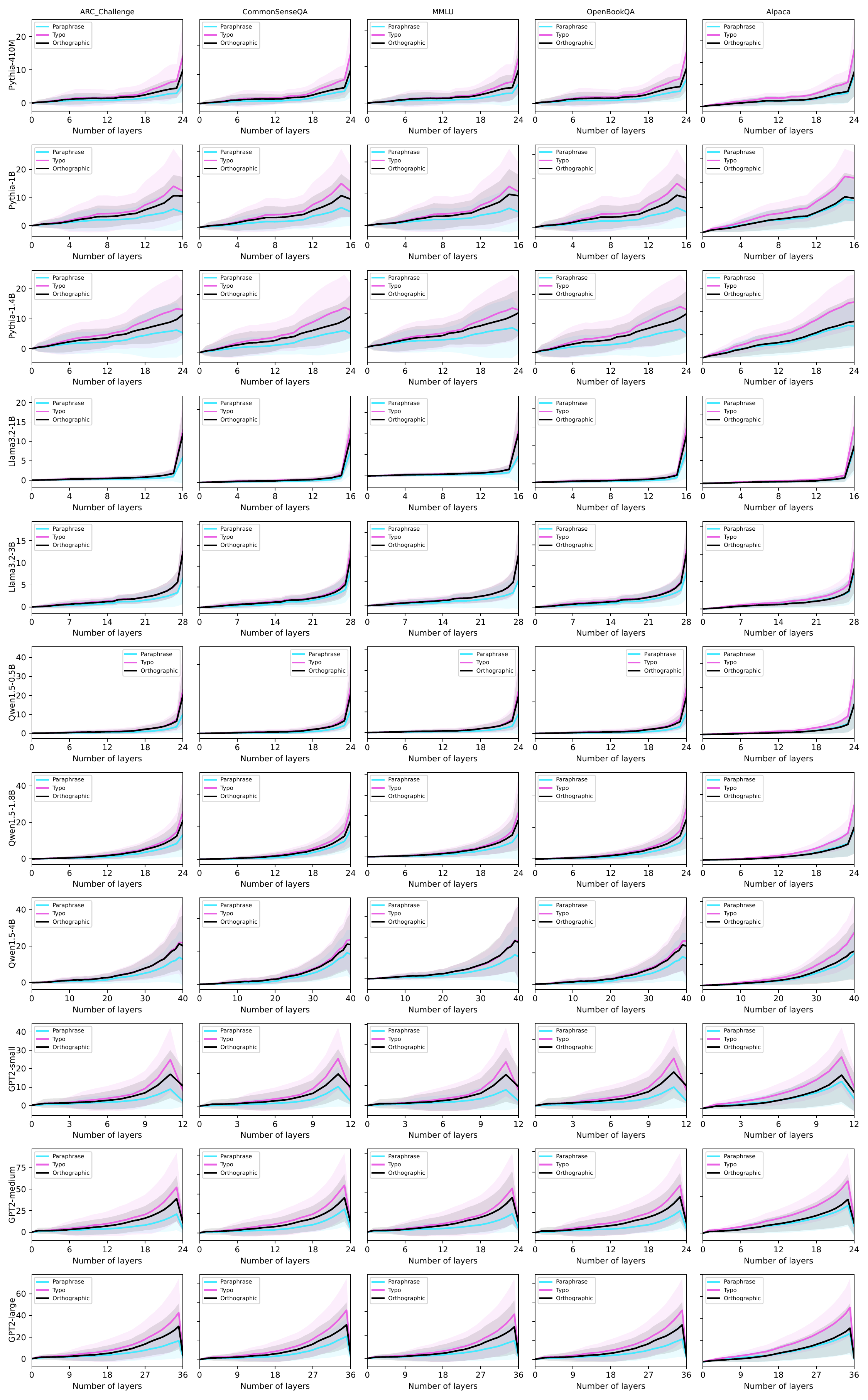}
  \caption{The comparison results between ``Paraphrase,'' ``Typo,'' and ``Orthographic'' across all models and datasets.}
  \label{fig:full_result_of_rq2_para_typo_orth}
\end{figure*}

In this section, we provide the results of all models across the four datasets.
Figure~\ref{fig:full_result_of_rq2_first_latter} compares ``First'' and ``Latter.''
We observe that the ordering between $\|\deltaH\|$ for ``First'' and ``Latter'' is not universal, but depends on the model series.
Within the \texttt{Pythia} series, $\|\deltaH\|$ for ``Latter'' is consistently higher than $\|\deltaH\|$ for ``First'' across all examined sizes (\texttt{Pythia-410M}, \texttt{Pythia-1B}, and \texttt{Pythia-1.4B}).
Within the \texttt{GPT2} series, the ordering varies inconsistently with model size: $\|\deltaH\|$ for ``Latter'' is lower than for ``First'' on \texttt{GPT2-small} and \texttt{GPT2-medium}, but higher on \texttt{GPT2-large}, so no monotonic size trend is observed.
Within the \texttt{Qwen1.5} series, we observe a clear size-dependent transition: $\|\deltaH\|$ for ``Latter'' is higher than for ``First'' on \texttt{Qwen1.5-0.5B} and \texttt{Qwen1.5-1.8B}, becomes comparable on intermediate sizes, and is exceeded by $\|\deltaH\|$ for ``First'' on \texttt{Qwen1.5-4B}.
Within the \texttt{Llama3.2} series, $\|\deltaH\|$ for ``Latter'' is higher than for ``First'' on \texttt{Llama3.2-1B}, and the two become comparable on \texttt{Llama3.2-3B}, which loosely follows the same size-dependent transition.
These results refine the earlier conclusion: the size-dependent transition from latter-sensitive to first-sensitive holds within more recent model series (\texttt{Qwen1.5} and \texttt{Llama3.2}), but does not extend to older series (\texttt{Pythia} and \texttt{GPT2}). We attribute this cross-series variation to differences in pretraining data and objectives across model eras.
Figure~\ref{fig:full_result_of_rq2_fewer_more} compares ``Fewer'' and ``More,'' in all cases, $\|\deltaH\|$ for ``More'' is higher than $\|\deltaH\|$ for ``Fewer.''
Figure~\ref{fig:full_result_of_rq2_modify_misalign} compares ``Modify'' and ``Misalign,'' in all cases, $\|\deltaH\|$ for ``Modify'' is higher than $\|\deltaH\|$ for ``Misalign.''
Figure~\ref{fig:full_result_of_rq2_para_typo_orth} compares ``Paraphrase,'' ``Typo,'' and ``Orthographic'' across all 11 models and 5 datasets.
In most cases, $\|\deltaH\|$ decreases in the order ``Typo'' > ``Orthographic'' > ``Paraphrase,'' consistent with the observation on Alpaca in Figure~\ref{fig:result_of_rq2_6}. 
All the experimental results align with the conclusions of RQ2.

\section{Mitigating Prompt Sensitivity via Activation Steering}
\label{sec:mitigating}

The Taylor-expansion analysis in \S~\ref{sec:interpretation_of_prompt_sensitivity} and experimental verification in \S~\ref{sec:experimental_verifications} identify $\|\deltaH\|$ as the primary driver of the upper bound on prompt sensitivity. This suggests that reducing $\|\deltaH\|$ at a target layer should directly reduce the model's output divergence between meaning-preserving prompts. We verify this hypothesis with activation steering, an intervention that forces $\|\deltaH^{(l)}\| = 0$ at a chosen layer $l$.

\subsection{Method}

Given a seed prompt $p_A$ and a meaning-preserving variant $p_B$, let $\h_A^{(l)}, \h_B^{(l)}$ denote their hidden states at layer $l$. We construct a steered forward pass for $p_A$ by overwriting $\h_A^{(l)} \leftarrow \h_B^{(l)}$ and continuing the forward computation with the original model weights for layers $l+1, \ldots, L$. This sets $\|\deltaH^{(l)}\| = 0$ at the intervention layer. We then measure the resulting prompt sensitivity as $|\deltaLogP|$ between the steered forward of $p_A$ and the natural forward of $p_B$.

\subsection{Experiments}

\begin{figure}[H]
  \includegraphics[width=0.9\columnwidth]{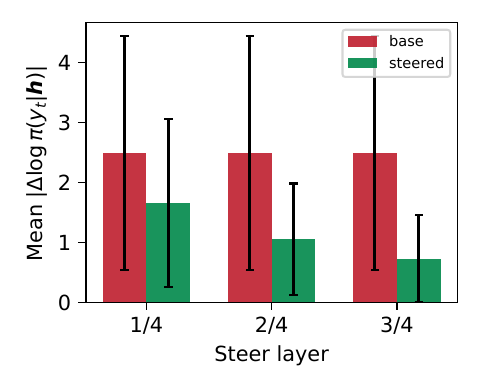}
  \caption{Activation steering on \texttt{Qwen1.5-4B} (ARC Challenge). Bars show the mean $|\deltaLogP|$ before (red) and after (green) steering at three layer depths $l \in \{L/4, L/2, 3L/4\}$.}
  \label{fig:steer_one}
\end{figure}

\begin{figure*}[htbp]
    \centering
    \includegraphics[width=0.70\textwidth]{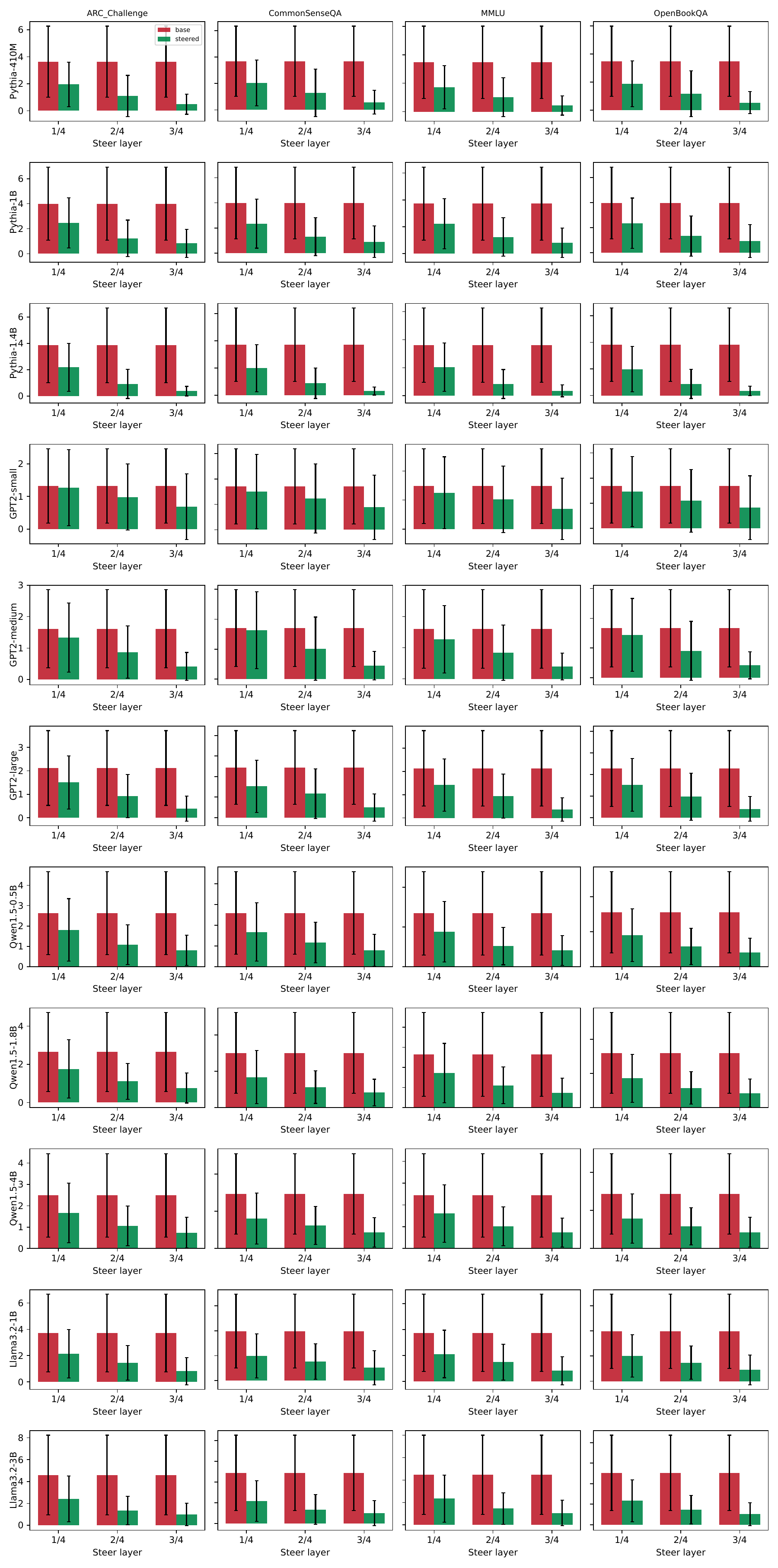} 
    \caption{Activation steering across all 11 models on the four MCQ datasets. Each cell shows $|\deltaLogP|$ before (red) and after (green) steering at $l \in \{L/4, L/2, 3L/4\}$.}
    \label{fig:steer_grid}
\end{figure*}

We apply steering at three depths $l \in \{L/4, L/2, 3L/4\}$ for all 11 models on the four MCQ datasets. Figure~\ref{fig:steer_one} reports $|\deltaLogP|$ averaged over meaning-preserving prompt pairs, comparing the non-steered baseline (red) with the steered forward pass (green). 
Steering consistently reduces prompt sensitivity, and the reduction grows as the intervention layer goes deeper. For example, on \texttt{Qwen1.5-4B} with the ARC Challenge, the baseline $|\deltaLogP|$ is $2.49$; steering at $L/4$, $L/2$, and $3L/4$ reduces it to $1.66$, $1.06$, and $0.73$ respectively. This confirms that $\|\deltaH\|$ plays a causal role in the observed prompt sensitivity predicted by our Taylor-expansion analysis: forcing $\|\deltaH\|$ to zero at any layer proportionally lowers the downstream log-probability divergence, with the effect being strongest when the intervention occurs closer to the output. Full results across the 11 models and 4 datasets are provided in Figure~\ref{fig:steer_grid}.
Across all combinations, steering reduces $|\deltaLogP|$ at every intervention depth, with monotonically larger reductions at deeper layers. This pattern further confirms that $\|\deltaH\|$ causally drives prompt sensitivity.

\section{Other Tokens as $y_t$ in Eq.~(\ref{eq:tailor_llm})}
\label{sec:tailor_llm_more_tokens}

\begin{figure*}[htbp]
    \centering
    \includegraphics[width=0.75\textwidth]{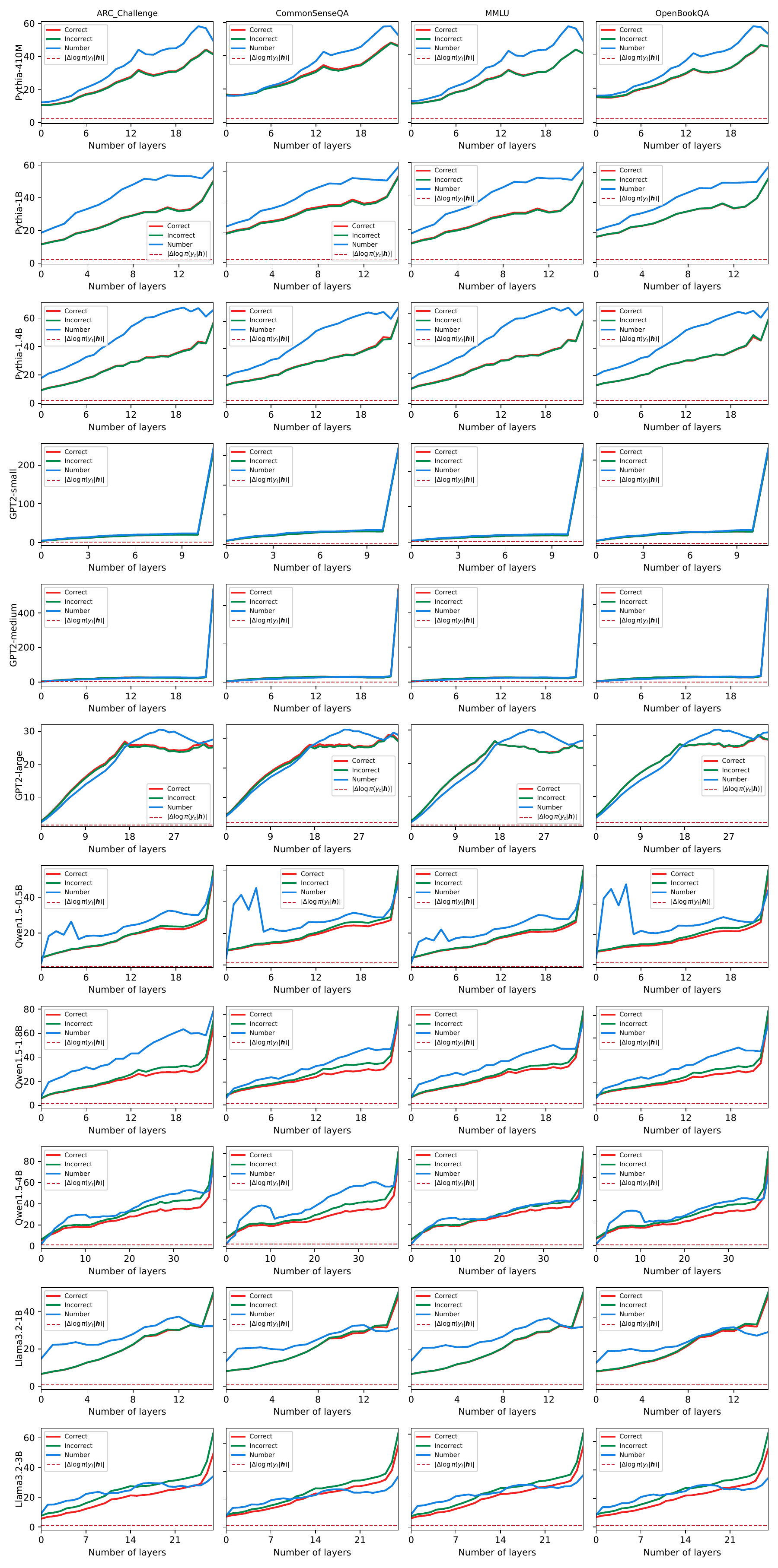} 
    \caption{The comparison of the upper bound $\upperBound$ of different $y_t$.}
    \label{fig:shuffle_choice_upper_bound}
\end{figure*}

\begin{figure*}[htbp]
    \centering
    \includegraphics[width=0.70\textwidth]{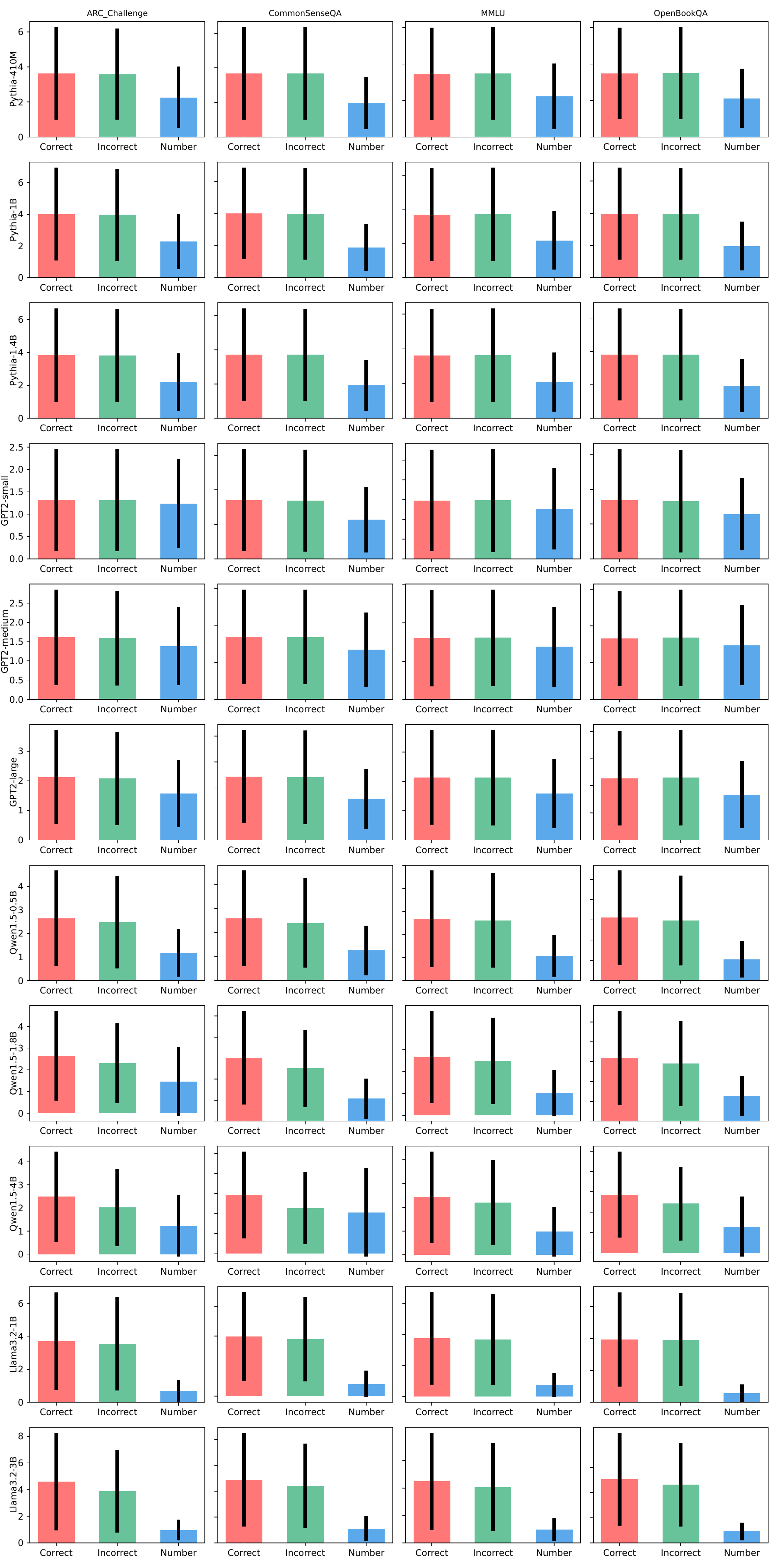} 
    \caption{The comparison of the absolute difference of the log probabilities $|\deltaLogP|$ of different $y_t$. ``Correct'', ``Incorrect'', and ``Number'' indicates the next token is $\yc$, $\yi$, and $\yn$, respectively.}
    \label{fig:shuffle_choice_delta_log_prob}
\end{figure*}

In the definition of Eq.~(\ref{eq:tailor_llm}), $y_t$ can be any token in the vocabulary of the LLMs.
This means that the logits of two meaning-preserving prompts should be equal at every position.
In this appendix, we analyze whether different values of $y_t$ affect the experimental results. Specifically, in addition to the correct answer, we randomly set $y_t$ to the following two types of values:
\begin{enumerate}
    \item $\yi$: A random sample of incorrect answers from the question's options.
    \item $\yn$: A random sample of numbers between 0 and 9.
\end{enumerate}

We mainly demonstrate the effects of the next token $y_t$ on the following two terms:

\begin{enumerate}
    \item The upper bound $\upperBound$ of different $y_t$.
    \item The absolute difference of the log probabilities $|\deltaLogP|$.
\end{enumerate}

Figure~\ref{fig:shuffle_choice_upper_bound} shows the comparison of the upper bounds for $\yc$, $\yi$, and $\yn$. 
We find that when $y_t$ is either the correct or incorrect option token, the trends of their upper bounds are similar.
The upper bound when $y_t$ is a number is higher than the upper bound when $y_t$ is an option (correct or incorrect).

As shown in Figure~\ref{fig:shuffle_choice_delta_log_prob}, we compare the absolute differences of the log probabilities for $\yc$, $\yi$, and $\yn$.
We can observe that when the next token is an option (correct or incorrect), the value of the absolute difference of the log probabilities is relatively close. However, when the next token is not an option, the value of the absolute difference of the log probabilities becomes significantly lower.
This is because the shape of the output probability distribution is significantly sharp, assigning higher probabilities to option tokens and lower probabilities to non-option tokens~\citep{liu2026alignment}.
Consequently, the difference between the two lower values has the chance to be relatively lower.
In summary, our analysis holds true for any next token $y_t$.


\end{document}